\pdfoutput=1

\documentclass[11pt]{article}

\usepackage[]{acl}

\usepackage{times}
\usepackage{latexsym}
\usepackage{multirow}
\usepackage{booktabs}
\usepackage{graphicx}
\usepackage{amssymb}
\usepackage{makecell}
\usepackage{amsmath, bm}
\usepackage{longtable}
\usepackage{subfigure}
\usepackage{caption}
\usepackage{enumerate}
\usepackage{enumitem}
\usepackage{longtable}
\usepackage{xspace}
\usepackage{comment}
\usepackage{cuted}
\usepackage{url}
\usepackage{amssymb}
\usepackage{pifont}

\usepackage{times}
\usepackage{helvet}
\usepackage{courier}
\usepackage{xcolor}
\usepackage{tcolorbox}
\usepackage{amsmath}

\usepackage{amsmath}
\usepackage{amssymb}
\usepackage{mathtools}
\usepackage{amsthm}
\usepackage{multirow}
\usepackage[capitalize,noabbrev]{cleveref}
\usepackage{fontawesome}

\usepackage{algorithm}
\usepackage{algpseudocode}
\usepackage{amsmath}

\usepackage[T1]{fontenc}

\usepackage[utf8]{inputenc}

\usepackage{microtype}

\usepackage{inconsolata}

\title{Controlling Multimodal Conversational Agents \\with Coverage-Enhanced Latent Actions}

\author{
\textbf{Yongqi Li$^{1,2,*,\dagger}$}, 
\textbf{Hao Lang$^{2,\dagger}$}, 
\textbf{Tieyun Qian$^{1,3,\ddag}$},
\textbf{Yongbin Li$^{2,\ddag}$}\\
        $^1$ School of Computer Science, Wuhan University, $^2$ Tongyi Lab\\
        $^3$ Zhongguancun Academy\\
        \small{\texttt{\{liyongqi,qty\}@whu.edu.cn}, \texttt{\{hao.lang,shuide.lyb\}@alibaba-inc.com}}
        }

\begin{document}
\maketitle

\renewcommand{\thefootnote}{\fnsymbol{footnote}}
\footnotetext[1]{{ }Work done while the author was interning at Tongyi Lab.}
\footnotetext[2]{{ }Equal contributions.}
\footnotetext[3]{{ }Corresponding authors.}
\renewcommand{\thefootnote}{\arabic{footnote}}

\begin{abstract}
Vision-language models are increasingly employed as multimodal conversational agents (MCAs) for diverse conversational tasks. 
Recently, reinforcement learning (RL) has been widely explored for adapting MCAs to various human-AI interaction scenarios. Despite showing great enhancement in generalization performance, fine-tuning MCAs via RL still faces challenges in handling the extremely large text token space. 
To address this, we learn a compact latent action space for RL fine-tuning instead. 
Specifically, we adopt the learning from observation mechanism to construct the codebook for the latent action space, where future observations are leveraged to estimate current latent actions that could further be used to reconstruct future observations.
However, the scarcity of paired image-text data hinders learning a codebook with sufficient coverage.
Thus, we leverage both paired image-text data and text-only data to construct the latent action space, using a cross-modal projector for transforming text embeddings into image-text embeddings.
We initialize the cross-modal projector on paired image-text data, and further train it on massive text-only data with a novel cycle consistency loss to enhance its robustness.
We show that our latent action based method outperforms competitive baselines on two conversation tasks across various RL algorithms. 
Code and data are available at \url{https://github.com/AlibabaResearch/DAMO-ConvAI/tree/main/MMLatentAction}.
\end{abstract}


\section{Introduction}
Vision-language models (VLMs)~\cite{yin-2024-MLLMsurvey} like Qwen-VL~\cite{bai-2025-qwen3vl} and GPT-4o~\cite{hurst-2024-gpt4ocard} are increasingly employed as multimodal conversational agents (MCAs) for various conversation tasks~\cite{yao-2025-MLLMAgentsurvey}.
MCAs enable emotionally rich and contextually grounded dialogues based on understanding both input images and texts, and thus become particularly valuable in fields like entertainment~\cite{mehta-2022-exploring}, online education~\cite{griol-2014-developing}, and personalized assistants~\cite{nguyen-2024-yo}.

Recently, reinforcement learning (RL)~\cite{sutton-1998-reinforcement} has been widely explored for adapting MCAs to diverse real-world human-AI interaction scenarios~\cite{zhou-2025-reinforcedMLLM}.
Generally, RL algorithms frame response token generation in MCAs as a sequential decision-making process~\cite{chen-2021-decision}, which optimize the policy to maximize cumulative rewards through interacting with environments.
Despite showing great enhancement in generalization performance~\cite{chu-2025-sftRL}, fine-tuning MCAs via RL still faces challenges in dealing with large exploration spaces.
For instance, with token vocabulary size $|\mathcal{V}|$ and maximum response length $m$, the sampling space for RL scales exponentially as $|\mathcal{V}|^m$.

To address the challenge of large text token space, we learn a compact latent action space for RL fine-tuning instead, following previous works~\cite{jia-2025-controlling}.
Specifically, we adopt the learning from observation mechanism~\cite{jiang-2023-efficient,ye-2025-latent} to construct the codebook for the latent action space, where future observations are leveraged to estimate current latent actions that could be further used to reconstruct future observations.
As a result, the action sampling space at each step is reduced from the token vocabulary size $|\mathcal{V}|$ (e.g., 152K for Qwen2.5-VL~\cite{bai-2025-qwen25vl}) to the latent action codebook size $|\mathcal{C}|$ (e.g., 128).

Generally, the codebook has to be learned from diverse data with sufficient coverage, which is a prerequisite for effective RL exploration in latent spaces~\cite{chen-2025-coverage}.
Note that VLMs in MCAs are typically pre-trained on paired image-text corpora $(V, T)$, which implicitly convey complementary and partially redundant information between visual and textual modalities~\cite{radford-2021-clip}.
Unfortunately, while unpaired image collections and text corpora are abundant on the web, curating them into aligned image-text corpora remains prohibitively costly~\cite{gupta-2025-better}, posing a dilemma in constructing latent spaces.
On one hand, using limited paired data and abundant unpaired data would introduce \textit{unimodal bias}~\cite{zhang-2024-UniModalBias}, where a model would overly rely on one modality and ignore others.
On the other hand, training the codebook solely on limited paired data may result in insufficient coverage, thereby impairing the agent’s generalization ability when handling diverse unseen conversation scenarios.

In this paper, we leverage both paired image-text data $(V,T)$ and unpaired text-only data $T$ to learn the codebook for the latent space.
To improve the coverage of latent actions while avoiding potentially unimodal bias, we attempt to construct pseudo paired data $(V^{\prime},T)$ based on text-only data $T$, and use the pseudo data $(V^{\prime},T)$ and the collected data $(V,T)$ to learn the codebook.

However, training a conditional image generator $G(V|T)$ for this purpose is computationally expensive due to the high dimension nature of images~\cite{pope-2021-dimension}.
Thus, we learn a cross-modal projector $P$ instead, which transforms an input text $e^T$ to an image-text pair $e^{V,T}$ in the embedding space, based on the cross-modal redundancy assumption~\cite{radford-2021-clip}.
Concretely, for each item in the paired image-text data $(V,T)$, we compute the text embedding $e^T$ and image-text embedding $e^{V,T}$ using an existing encoder, and train the projector $P$ to imitate the projection between these two kinds of embeddings.
To enhance the robustness of the projector $P$, we further train it on massive text-only data $T$ using a cycle consistency loss~\cite{zhu-2017-unpaired}.
We introduce an additional projector $P^{\prime}$ that can transform image-text embedding $e^{V,T}$ back to text embedding $e^T$.
In this way, we can optimize the projector $P$ by enforcing cycle consistency on text-only data $T$ such that $P^{\prime}(P(e^{T}))\approx e^{T}$.

We evaluate our method on two conversation tasks, namely multimodal role-playing conversation~\cite{dai-2025-mmrole} and multimodal personalized conversation~\cite{li-2025-aligning}.
To evaluate the generalizability of latent actions, we conduct experiments using various RL algorithms, such as GRPO~\cite{shao-2024-GRPO} and Dr.GRPO~\cite{liu-2025-DRGRPO}.
We construct the latent action space using paired image-text data $(V, T)$ and text-only data $T$.
The $(V, T)$ data are comprised of image-caption pairs, multimodal news articles, and multimodal Wikipedia pages, totaling 14M images and 1B text tokens.
The text-only data are mainly derived from SlimPajama~\cite{cerebras-2023-slimpajama}, which contains 627B text tokens.  
Experimental results show that our method outperforms competitive baselines.

In summary, our work makes the following three key contributions.
1) We are the first to introduce latent actions for fine-tuning multimodal conversational agents via RL, which significantly reduces the exploration space.
2) We construct the latent action space with both paired image-text data and text-only data, using a cross-modal projector trained with a novel cycle consistency loss.
3) We evaluate our latent action based method on two multimodal conversation tasks and demonstrate that our method outperforms competitive baselines, and further show that the cross-modal projector is critical for improving the coverage of latent actions.

\section{Preliminary}

\paragraph{Reinforcement Learning for VLM Agents}
In reinforcement learning (RL), problems are framed by a Markov Decision Process (MDP) \(\mathcal{M} = \langle \mathcal{S}, \mathcal{A}, \mathcal{T}, \mathcal{R} \rangle\). 
For VLMs, the state at step \(t\) is the contextual information \(s_t = (x^V, x^{T_{1:t}}) \in \mathcal{S}\), which includes the input image \(x^V\) and the current token sequence \(x^{T_{1:t}}\).
\(\mathcal{A}\) is the action space containing all possible actions \(a_t\) at each step.
\(\mathcal{T}\) is the state transition function, governing the transition from \(s_t\) to \(s_{t+1}\), i.e., \(P\big(s_{t+1} \mid s_t, a_t\big)\).
The reward function $\mathcal{R}\big(x^{T_{p+1:m}}\big)$ assigns a scalar reward to the response $x^{T_{p+1:m}}$, conditioned on the input $(x^V, x^{T_{1:p}})$, with prompt length $p$ and maximum sequence length $m$, following common practice in RL for VLMs~\cite{shen-2025-vlmR1}.


\paragraph{Latent Actions for Reinforcement Learning}
In traditional token-level RL, each action \(a_t\) corresponds to selecting the next text token \(x^{T_{t+1}}\) from the token vocabulary \(\mathcal{V}\), i.e., $\mathcal{A}=\mathcal{V}$. While in latent action RL, at each step $t$, the policy \(\pi_\theta(a_t | x^V, x^{T_{1:t}})\) selects a latent action \(a_t\) from a compact codebook \(\mathcal{C}\), i.e., $\mathcal{A}=\mathcal{C}$.
During RL exploration, the latent action policy samples a latent action at each step, ultimately yielding the terminal state \(s_m\). 
During exploitation, the latent action policy is refined to maximize expected rewards using RL algorithms such as GRPO~\cite{shao-2024-GRPO}.

\begin{figure}[t]
\centering
\includegraphics[width=0.45\textwidth]{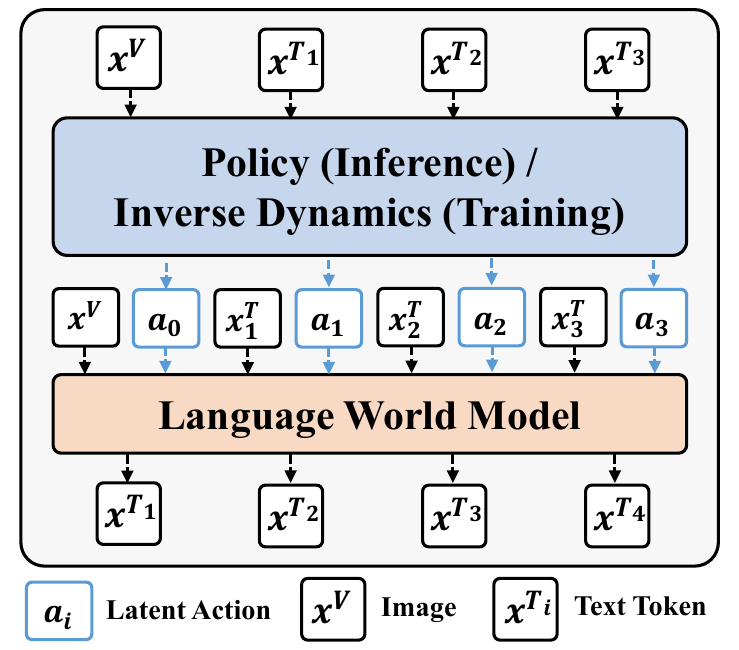}
\caption{Illustrations of integrating latent actions with vision-language models.}
\label{fig:model_design}
\end{figure}

\begin{figure*}[t]
\centering
\includegraphics[width=0.9\textwidth]{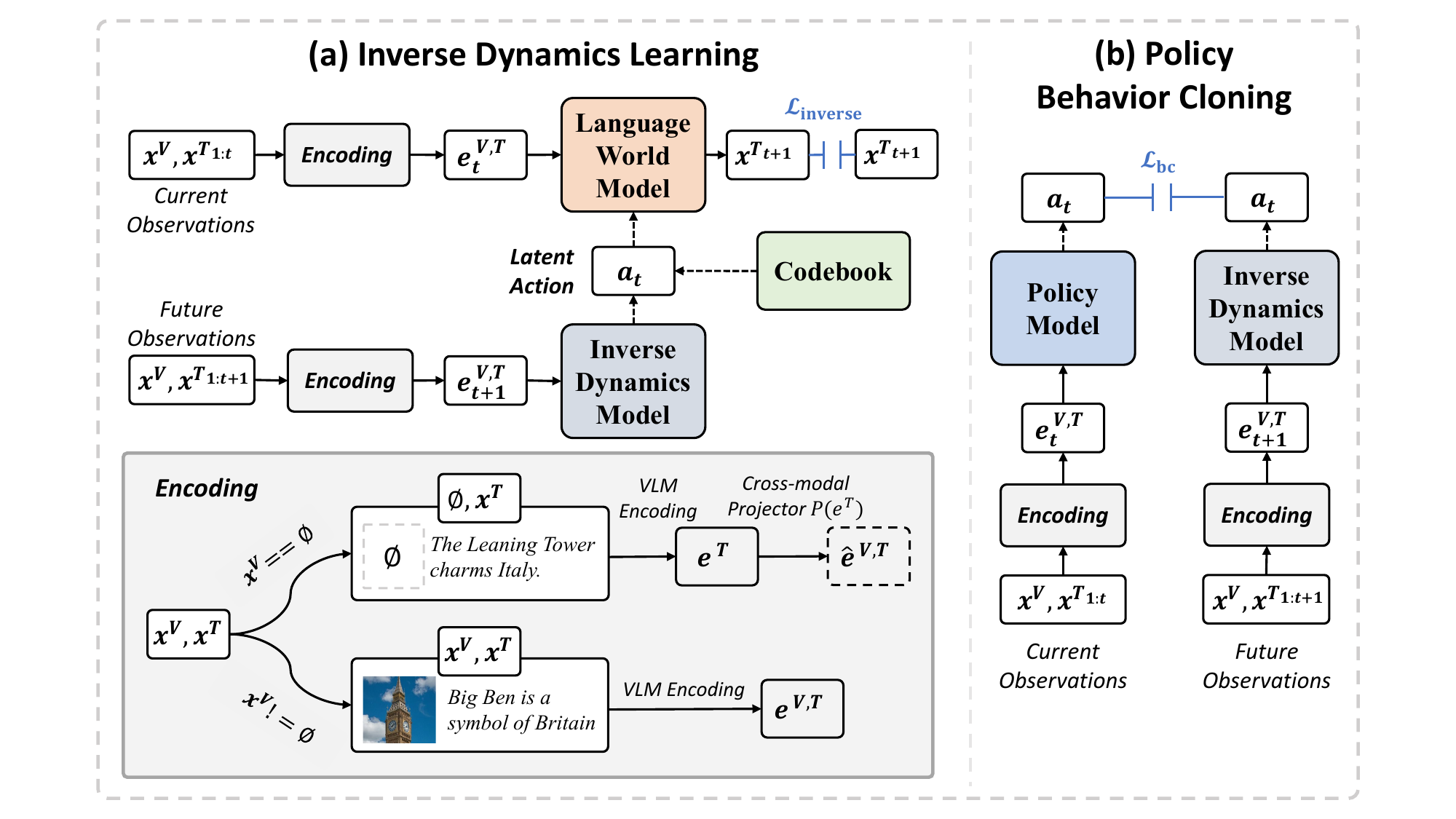}
\caption{Pipeline for constructing the latent action space. (a) \textbf{Inverse dynamics learning}: Given future observations, the inverse dynamics model infers a discrete latent action from a learnable codebook; the language world model then uses this latent action and current observations to reconstruct the next token $x^{T_{t+1}}$. The language world model, inverse dynamics model, and codebook are jointly trained. (b) \textbf{Policy behavior cloning}: A policy model is trained to predict the same latent actions as those inferred by the inverse dynamics model, using only current observations.}
\label{fig:framework}
\end{figure*}

\section{Methodology}
In this section, we first describe the overall model design for incorporating latent actions into VLMs (Sec.~\ref{sec:method:model}).
Next, we detail the unsupervised construction of the latent action space (Sec.~\ref{sec:method:inverse_and_bc}).
Finally, we introduce the procedure of latent action based RL fine-tuning (Sec.~\ref{sec:method:RL}).

\subsection{Model Design}\label{sec:method:model}
To fine-tune MCAs via latent action RL, we introduce three new modules, as illustrated in Figure~\ref{fig:model_design}. These modules are designed to share a base VLM while adding a small number of additional parameters, thereby introducing only marginal computational overhead. For further details on the model design, please refer to the Appendix~\ref{sec:app:detail_model_design}.

\paragraph{Language World Model $f_{\text{world}}$} The language world model $f_{\text{world}}(x^{T_{t+1}}|x^V,x^{T_{1:t}},a_t)$ takes current observations \((x^V, x^{T_{1:t}})\) and a latent action \(a_t\) as input, and auto-regressively outputs the next token \(x^{T_{t+1}}\).
The latent action \(a_t\) is provided by the inverse dynamics model \(f_{\text{inverse}}\) during constructing the latent action space, and by the policy \(\pi_\theta\) during inference and RL phases.

\paragraph{Inverse Dynamics Model $f_{\text{inverse}}$} The inverse dynamics model $f_{\text{inverse}}(a_t | x^V, x^{T_{1:t+1}})$ takes future observations \((x^V, x^{T_{1:t+1}})\) as input, and outputs a discrete latent action index \(a_t \in \{1, \dots, |\mathcal{C}|\}\) for the current step.
The corresponding latent action embedding \(c_{a_t} = \mathcal{C}[a_t] \in \mathbb{R}^d\) is then retrieved from the trainable codebook \(\mathcal{C} \in \mathbb{R}^{|\mathcal{C}| \times d}\) and used by \(f_{\text{world}}\) to reconstruct the next token \(x^{T_{t+1}}\).
Note that $f_{\text{inverse}}$ only assists training and does not serve for the inference phase.

\paragraph{Policy Model $\pi_\theta$}
The latent action policy model $\pi_\theta(a_t | x^V, x^{T_{1:t}})$ takes the current observations \((x^V, x^{T_{1:t}})\) as input, and predicts latent action $a_t$ for the current step.
Since the language world model $f_{\text{world}}$ is controlled by latent actions, we can optimize the latent action distribution of $\pi_\theta$ for steering $f_{\text{world}}$ to generate responses toward higher rewards.

\subsection{Latent Action Space Learning}\label{sec:method:inverse_and_bc}
Following~\citet{jia-2025-controlling}, we construct the latent action space using large-scale corpora in two steps. 1) \textit{inverse dynamics learning}, which trains the $f_{\text{world}}$, $f_{\text{inverse}}$, and $\mathcal{C}$ in an unsupervised manner (Fig.~\ref{fig:framework} (a)); 2) \textit{policy behavior cloning}, which trains the policy model $\pi_\theta$ to mimic the latent action \(a_t\) inferred by \(f_{\text{inverse}}\) (Fig.~\ref{fig:framework} (b)).

\subsubsection{Inverse Dynamics Learning}
We first outline the overall objective of inverse dynamics learning, followed by the training procedure of the introduced cross-modal projector.

\paragraph{Overview}
As shown in Fig.~\ref{fig:framework} (a), we jointly train the inverse dynamics model \( f_{\text{inverse}} \), language world model \( f_{\text{world}} \), and the latent action codebook \( \mathcal{C} \), using the mixed corpus \( \mathcal{D}^{VT} \cup \mathcal{D}^{T} \) (paired image-text data and text-only data).
The loss is as:
\begin{equation}\label{eq:inverse_learning}
\footnotesize
\mathcal{L}_{\text{inverse}} = \mathbb{E}_{\mathcal{D}^{VT} \cup \mathcal{D}^{T}} \left[ - \sum_{t=1}^{m-1} \log f_{\text{world}} \big( x^{T_{t+1}} | e^{V,T}_t, a_t \big) \right],
\end{equation}
where the expectation is taken over sequences \((x^V, x^{T_{1:m}})\) sampled from the mixed corpus \(\mathcal{D}^{VT} \cup \mathcal{D}^{T}\), with \(a_t = f_{\text{inverse}}(e^{V,T}_{t+1}) \in \{1,...,|\mathcal{C}|\}\).
The embedding \( e^{V,T}_t \) is obtained via:
\begin{equation}\label{eq:e_vt}
\footnotesize
e^{V,T}_t = 
\begin{cases}
f_{\text{VLM}}(x^V, x^{T_{1:t}}), & \text{if } x^V \neq \emptyset \quad (\text{from } \mathcal{D}^{VT}); \\
P\big( f_{\text{VLM}}(x^{T_{1:t}}) \big), & \text{if } x^V = \emptyset \quad (\text{from } \mathcal{D}^{T}),
\end{cases}
\end{equation}
where \( f_{\text{VLM}} \) denotes the encoding module based on VLMs. 
$P$ denotes the cross-modal projector for transforming text embeddings into image-text embeddings, and its training procedure is as follows.

\paragraph{Cross-modal Projector Training}
Let \(P\) denote the forward cross-modal projector, which maps text embeddings \(e^T_t\) to the parameters of a diagonal Gaussian distribution over the image-text embedding space, i.e., \((\mu_{t}, \sigma_{t}) = P(e^T_t)\).
Let \(P'\) denote the reverse projector, which maps image-text embeddings back to the text embedding space.
We train \(P\) and \(P'\) in the following two steps.

\textbf{\textit{Step 1: Initialization on paired image-text data.}}
We first train the forward projector \(P\) on paired image-text data \(\mathcal{D}^{VT}\), where the loss is defined as:
\begin{equation}\label{eq:proj_t2vt}
\footnotesize
\mathcal{L}_{\text{t2vt}} = \mathbb{E}_{\mathcal{D}^{VT}} \left[ 
    \sum_{t=1}^{m-1} \frac{1}{2} \left( 
        \left\| \frac{e^{V,T}_{t} - \mu_{t}}{\sigma_{t}} \right\|^2 
        + \|\log \sigma_{t}^2\|_1
    \right)
\right],
\end{equation}
where the expectation is taken over sequences \((x^V, x^{T_{1:m}}) \sim \mathcal{D}^{VT}\),  
and \(e^{V,T}_{t} = f_{\text{VLM}}(x^V, x^{T_{1:t}})\), and \((\mu_{t}, \sigma_{t})= P(e^T_t = f_{\text{VLM}}(x^{T_{1:t}}))\).

Similarly, $P'$ is trained on $\mathcal{D}^{VT}$ using the symmetric loss $\mathcal{L}_{\text{vt2t}}$, defined as:
\begin{equation}\label{eq:proj_vt2t}
\footnotesize
\mathcal{L}_{\text{vt2t}} = \mathbb{E}_{\mathcal{D}^{VT}} \left[ 
    \sum_{t=1}^{m-1} \frac{1}{2} \left( 
        \left\| \frac{e^{T}_{t} - \nu_{t}}{\tau_{t}} \right\|^2 
        + \|\log \tau_{t}^2\|_1 
    \right)
\right],
\end{equation}
where the expectation is taken over sequences $(x^V, x^{T_{1:m}}) \sim \mathcal{D}^{VT}$,  
$e^{T}_{t} = f_{\text{VLM}}(x^{T_{1:t}})$ denotes the text embedding,  
and $(\nu_{t}, \tau_{t}) = P'(e^{V,T}_{t} = f_{\text{VLM}}(x^V, x^{T_{1:t}}))$.  
The total loss for \textit{Step 1} is:
\begin{equation}\label{eq:proj_step1_total}
    \mathcal{L}_{\text{proj}_\text{1}} = \mathcal{L}_{\text{t2vt}} + \mathcal{L}_{\text{vt2t}}.
\end{equation}

\textbf{\textit{Step 2: Jointly training on paired image-text data and text-only data}}
We now jointly train \(P\) and \(P'\) on paired data \(\mathcal{D}^{VT}\) and text-only data \(\mathcal{D}^T\). The total objective is: 
\begin{equation}\label{eq:proj_step2_total}
\mathcal{L}_{\text{proj}_\text{2}} = 
\mathcal{L}_{\text{t2vt}} + 
\mathcal{L}_{\text{vt2t}} + 
\mathcal{L}_{\text{cycle}}
\end{equation}
where \(\mathcal{L}_{\text{t2vt}}\) (Eq.~\ref{eq:proj_t2vt}) and \(\mathcal{L}_{\text{vt2t}}\) (Eq.~\ref{eq:proj_vt2t}) are computed over \(\mathcal{D}^{VT}\), and \(\mathcal{L}_{\text{cycle}}\) denotes a novel cycle consistency loss computed on text-only data \(\mathcal{D}^T\).

The cycle consistency loss $\mathcal{L}_{\text{cycle}}$ is defined as:
\begin{equation}\label{eq:cycle}
\footnotesize
\mathcal{L}_{\text{cycle}} = \mathbb{E}_{\mathcal{D}^{T}} \left[ 
    \sum_{t=1}^{m-1} \frac{1}{2} \left( 
        \left\| \frac{e^{T}_{t} - \nu_{t}}{\tau_{t}} \right\|^2 
        + \|\log \tau_{t}^2\|_1 
    \right)
\right],
\end{equation}
where the expectation is taken over text-only sequences $x^{T_{1:m}} \sim \mathcal{D}^{T}$,  
$e^{T}_{t} = f_{\text{VLM}}(x^{T_{1:t}})$, and $(\mu_{t}, \sigma_{t}) = P(e^{T}_{t})$, and $(\nu_{t}, \tau_{t}) = P'(\mu_{t})$.

\subsubsection{Policy Behavior Cloning}
During RL exploration and inference, future observations are unavailable, making the inverse dynamics model \(f_{\text{inverse}}\) inapplicable. 
Thus, we train a policy model \(\pi_\theta\) via behavior cloning to mimic latent actions inferred by \(f_{\text{inverse}}\) (Fig.~\ref{fig:framework} (b)).
Specifically, for samples from the mixed corpus $\mathcal{D}^{\text{mix}} = \mathcal{D}^{VT} \cup \mathcal{D}^{T}$, 
we compute the loss as:
\begin{equation}\label{eq:bc}
\footnotesize
\mathcal{L}_{\text{bc}} = \mathbb{E}_{\mathcal{D}^{\text{mix}}} \left[ 
    -\sum_{t=1}^{m-1} \log \pi_\theta \big(a_t^\ast = f_{\text{inverse}}(e^{V,T}_{t+1}) \mid e^{V,T}_t\big) 
\right],
\end{equation}
where the expectation is taken over sequences $(x^V, x^{T_{1:m}}) \sim \mathcal{D}^{\text{mix}}$,  
with $e^{V,T}_t$ defined as in Eq.~\ref{eq:e_vt}.

\subsection{Latent Action Reinforcement Learning}\label{sec:method:RL}
\begin{figure}[t]
\centering
\includegraphics[width=0.48\textwidth]{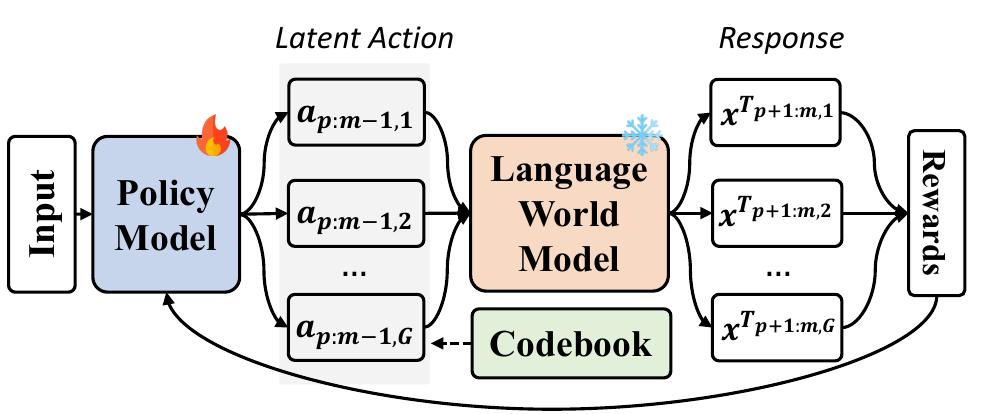}
\caption{Illustrations of latent action RL. The language world model is frozen, while the policy model is optimized to select latent actions from the codebook that steer the generated responses toward higher rewards.} 
\label{fig:latent_action_RL}
\end{figure}

On downstream multimodal conversational tasks, we perform reinforcement learning at the policy model level, as illustrated in Fig.~\ref{fig:latent_action_RL}.
For each prompt $(x^V, x^{T_{1:p}}) \sim \mathcal{D}_{\text{rl}}$ with the prompt length $p$, the policy $\pi_\theta$ and the world model $f_{\text{world}}$ jointly generate response $x^{T_{p+1:m}}$ auto-regressively, i.e., at each step $t = p, \dots, m-1$, $a_t \sim \pi_\theta(\cdot | x^V, x^{T_{1:t}}), x^{T_{t+1}} = f_{\text{world}}(x^V, x^{T_{1:t}}, a_t)$, with maximum length $m$. 
We optimize $\pi_\theta$ by maximizing the expected rewards:
\begin{equation}\label{eq:RL_objective}
\footnotesize
    \mathcal{J}(\theta) = \mathbb{E}_{(x^V, x^{T_{1:p}}) \sim \mathcal{D}_{\text{rl}}} \left[ R\big(x^{T_{p+1:m}}\big) \right],
\end{equation}
where $R(\cdot)$ denotes the reward function.
During latent action RL, we only optimize the latent action prediction distribution of the policy model while keeping the language world model (responsible for text token generation) frozen. Consequently, the length of the optimized sequence (latent action sequence) is equal to the response length, which means that our approach does not increase the length of the optimized sequence compared to token-level RL methods.

We summarize our framework in Algorithm~\ref{alg:latent_action}.

\begin{algorithm}[t]
\caption{Latent Action Space Learning and Latent Action RL}
\label{alg:latent_action}
\begin{algorithmic}[1]
\Statex \textbf{Stage 1: Latent Action Space Learning}
\State Initialize $f_{\text{world}},f_{\text{inverse}},\mathcal{C}$ by minimizing $\mathcal{L}_{\text{inverse}}$ (Eq.~\ref{eq:inverse_learning}) on $\mathcal{D}^{VT}$.
\State Initialize the cross-modal projectors $P, P'$ by minimizing $\mathcal{L}_{\text{proj}_1}$ (Eq.~\ref{eq:proj_step1_total}) on $\mathcal{D}^{VT}$.
\State Jointly optimize $f_{\text{world}},f_{\text{inverse}},\mathcal{C},P, P'$ by minimizing $\mathcal{L}_{\text{inverse}}$ (Eq.~\ref{eq:inverse_learning}) and $\mathcal{L}_{\text{proj}_2}$ (Eq.~\ref{eq:proj_step2_total}) on $\mathcal{D}^{VT} \cup \mathcal{D}^{T}$.
\State Initialize the policy model $\pi_\theta$ by minimizing $\mathcal{L}_{\text{bc}}$ (Eq.~\ref{eq:bc}) on $\mathcal{D}^{VT} \cup \mathcal{D}^{T}$.

\Statex \textbf{Stage 2: Latent Action RL}
\State Sample $(x^V, x^{T_{1:p}}) \sim \mathcal{D}_{\text{rl}}$:
\State \quad Roll out $x^{T_{p+1:m}}$ via $a_t \sim \pi_\theta(\cdot | x^V, x^{T_{1:t}})$, $x^{T_{t+1}} = f_{\text{world}}(x^V, x^{T_{1:t}}, a_t)$, $t=p,..., m-1$.
\State \quad Compute reward $R(x^{T_{p+1:m}})$.
\State \quad Optimize $\pi_\theta$ by maximizing $\mathcal{J}(\theta)$ (Eq.~\ref{eq:RL_objective}).
\end{algorithmic}
\end{algorithm}

\section{Experiments}
\subsection{Experimental Setup}
\paragraph{Models}
We build the language world model, inverse dynamics model, and policy model upon the same foundation vision-language model. Specifically, we use \texttt{Qwen2.5-VL-3B-Instruct} and \texttt{Qwen2.5-VL-7B-Instruct}~\cite{bai-2025-qwen25vl} for main experiments. 
The latent action space is implemented as a codebook with size \(|\mathcal{C}| = 128\).

\paragraph{Datasets}
During the latent action space construction stage (Section~\ref{sec:method:inverse_and_bc}), we use a mixture of paired image-text corpora \(\mathcal{D}^{VT}\) and text-only corpora \(\mathcal{D}^{T}\). 
For \(\mathcal{D}^{VT}\), we collect image-caption pairs from \texttt{Conceptual-12M}~\cite{changpinyo-2021-Conceptual}, multimodal news articles from \texttt{N24News}~\cite{wang-2022-N24News}, and multimodal Wikipedia data from \texttt{WikiWeb2M}~\cite{burns-2023-wiki}, totaling 14 million images and 1 billion text tokens.
For \(\mathcal{D}^{T}\), we collect text-only data mainly from the \texttt{SlimPajama-627B} dataset~\cite{cerebras-2023-slimpajama}, which contains 627 billion text tokens.

For latent action RL (Section~\ref{sec:method:RL}), we evaluate our method on two downstream tasks: 1) multimodal role-playing conversation on \texttt{MMRole}~\cite{dai-2025-mmrole}, where we focus on the challenging \texttt{Comment} subset; we train on the in-distribution (ID) split and evaluate on ID and out-of-distribution (OOD) test sets; 2) multimodal personalized conversation on \texttt{PCogAlignBench}~\cite{li-2025-aligning}, where we train the agent on the \texttt{LS1} set and evaluate on \texttt{LS1} and \texttt{LS2} test sets.

We select MMRole and PCogAlignBench because their responses average $\approx 200$ tokens (Figure~\ref{fig:app_case_study_MMRole} and~\ref{fig:app_case_study_PCogAlign}), yielding a significantly larger RL sampling space than traditional conversational benchmarks~\cite{haber-2019-photobook,de-2017-guesswhat,das-2017-visual}, whose responses are typically less than 20 tokens. Since the RL sampling space grows exponentially with response length~\cite{feng-2025-towards}, these benchmarks pose substantially greater challenges for RL fine-tuning, making them more suitable for evaluating our method.



\paragraph{Evaluation Metrics}
We adopt the \textit{LLM-as-a-Judge} metric to evaluate model performance, using prompt templates validated by \citet{dai-2025-mmrole,li-2025-aligning}, which show high correlation with human judgments.
For each sample, the LLM judge scores both the model and ground-truth responses across benchmark-specific dimensions, with scores ranging 1-10. 
Then, following~\cite{dai-2025-mmrole}, we report the ratio of the model’s average score to the ground-truth response’s average score across all evaluation dimensions.
We report the mean and standard deviation across three evaluation runs.

\begin{table*}[h]
\setlength{\tabcolsep}{2.6mm}
    \resizebox{\linewidth}{!}{
    \begin{tabular}{clccccc}
        \toprule
        \multicolumn{2}{c}{\multirow{2}{*}{\textbf{Method}}} 
        & \multicolumn{2}{c}{\textbf{MMRole}} 
        & \multicolumn{2}{c}{\textbf{PCogAlignBench}} 
        & \multirow{2}{*}{\textbf{Average}} \\
        \cmidrule(lr){3-4}\cmidrule(lr){5-6}
        & & \multicolumn{1}{c}{\textit{ID}} & \multicolumn{1}{c}{\textit{OOD}} 
        & \multicolumn{1}{c}{\textit{LS1}} & \multicolumn{1}{c}{\textit{LS2}} & \\
        \midrule

\multirow{10}{*}{\rotatebox{90}{\textit{Qwen2.5-VL-3B-Instruct}}} 
& Prompt          &0.728$_{\pm 0.005}$ & 0.687$_{\pm 0.025}$ & 0.678$_{\pm 0.003}$ & 0.676$_{\pm 0.002}$ & 0.692$_{\pm 0.009}$\\
& SFT             &0.843$_{\pm 0.002}$ & 0.809$_{\pm 0.012}$ & 0.808$_{\pm 0.009}$ & 0.810$_{\pm 0.005}$ & 0.817$_{\pm 0.007}$\\
\cmidrule(lr){2-7}
& GRPO (Token)    &0.838$_{\pm 0.017}$ & 0.796$_{\pm 0.027}$ & 0.845$_{\pm 0.007}$ & \textbf{0.845}$_{\pm 0.004}$ & 0.831$_{\pm 0.014}$\\
& GRPO (Latent Action)   &\textbf{0.949}$_{\pm 0.007}$ & \textbf{0.915}$_{\pm 0.065}$ & \textbf{0.871}$_{\pm 0.011}$ & 0.837$_{\pm 0.010}$ & \textbf{0.893}$_{\pm 0.023}$\\
\cmidrule(lr){2-7}
& Dr.GRPO (Token) &0.867$_{\pm 0.011}$ & 0.823$_{\pm 0.002}$ & 0.835$_{\pm 0.008}$ & 0.834$_{\pm 0.012}$ & 0.840$_{\pm 0.008}$\\
& Dr.GRPO (Latent Action)&\textbf{0.953}$_{\pm 0.016}$ & \textbf{0.916}$_{\pm 0.038}$ & \textbf{0.874}$_{\pm 0.009}$ & \textbf{0.840}$_{\pm 0.009}$ & \textbf{0.896}$_{\pm 0.018}$\\
\cmidrule(lr){2-7}
& DAPO (Token)    &0.856$_{\pm 0.003}$ & 0.805$_{\pm 0.033}$ & 0.835$_{\pm 0.008}$ & 0.828$_{\pm 0.008}$ & 0.831$_{\pm 0.013}$\\
& DAPO (Latent Action)   &\textbf{0.941}$_{\pm 0.016}$ & \textbf{0.889}$_{\pm 0.009}$ & \textbf{0.879}$_{\pm 0.011}$ & \textbf{0.835}$_{\pm 0.006}$ & \textbf{0.886}$_{\pm 0.010}$\\
\cmidrule(lr){2-7}
& BNPO (Token)    &0.860$_{\pm 0.012}$ & 0.801$_{\pm 0.038}$ & 0.849$_{\pm 0.008}$ & 0.836$_{\pm 0.007}$ & 0.836$_{\pm 0.016}$\\
& BNPO (Latent Action)   &\textbf{0.940}$_{\pm 0.004}$ & \textbf{0.901}$_{\pm 0.014}$ & \textbf{0.872}$_{\pm 0.007}$ & \textbf{0.836}$_{\pm 0.008}$ & \textbf{0.887}$_{\pm 0.008}$\\

\midrule
\midrule

\multirow{10}{*}{\rotatebox{90}{\textit{Qwen2.5-VL-7B-Instruct}}} 
& Prompt          &0.839$_{\pm 0.006}$ & 0.821$_{\pm 0.024}$ & 0.721$_{\pm 0.003}$ & 0.710$_{\pm 0.003}$ & 0.773$_{\pm 0.009}$\\
& SFT             &0.885$_{\pm 0.003}$ & 0.856$_{\pm 0.013}$ & 0.808$_{\pm 0.005}$ & 0.799$_{\pm 0.004}$ & 0.837$_{\pm 0.006}$\\
\cmidrule(lr){2-7}
& GRPO (Token)    &0.892$_{\pm 0.004}$ & 0.840$_{\pm 0.014}$ & 0.870$_{\pm 0.016}$ & 0.851$_{\pm 0.012}$ & 0.863$_{\pm 0.011}$\\
& GRPO (Latent Action)   &\textbf{0.920}$_{\pm 0.005}$ & \textbf{0.872}$_{\pm 0.016}$ & \textbf{0.898}$_{\pm 0.009}$ & \textbf{0.852}$_{\pm 0.010}$ & \textbf{0.885}$_{\pm 0.010}$\\
\cmidrule(lr){2-7}
& Dr.GRPO (Token) &0.892$_{\pm 0.006}$ & 0.854$_{\pm 0.009}$ & 0.854$_{\pm 0.006}$ & 0.839$_{\pm 0.004}$ & 0.860$_{\pm 0.006}$\\
& Dr.GRPO (Latent Action)&\textbf{0.916}$_{\pm 0.010}$ & \textbf{0.864}$_{\pm 0.020}$ & \textbf{0.897}$_{\pm 0.008}$ & \textbf{0.851}$_{\pm 0.015}$ & \textbf{0.882}$_{\pm 0.013}$\\
\cmidrule(lr){2-7}
& DAPO (Token)    &0.892$_{\pm 0.004}$ & 0.842$_{\pm 0.025}$ & 0.844$_{\pm 0.013}$ & 0.828$_{\pm 0.007}$ & 0.852$_{\pm 0.012}$\\
& DAPO (Latent Action)   &\textbf{0.920}$_{\pm 0.009}$ & \textbf{0.863}$_{\pm 0.017}$ & \textbf{0.903}$_{\pm 0.012}$ & \textbf{0.850}$_{\pm 0.005}$ & \textbf{0.884}$_{\pm 0.011}$\\
\cmidrule(lr){2-7}
& BNPO (Token)    &0.894$_{\pm 0.004}$ & \textbf{0.859}$_{\pm 0.029}$ & 0.850$_{\pm 0.007}$ & 0.836$_{\pm 0.004}$ & 0.860$_{\pm 0.011}$\\
& BNPO (Latent Action)   &\textbf{0.916}$_{\pm 0.006}$ & 0.842$_{\pm 0.018}$ & \textbf{0.901}$_{\pm 0.009}$ & \textbf{0.852}$_{\pm 0.012}$ & \textbf{0.878}$_{\pm 0.011}$\\
        \bottomrule
    \end{tabular}
    }
    \caption{Performance comparison on MMRole and PCogAlignBench, using the \textit{LLM-as-a-Judge} metric. Results are averaged over three runs. We conduct experiments using various VLMs, including Qwen2.5-VL-3B-Instruct and Qwen2.5-VL-7B-Instruct. Best results are in \textbf{bold} on each RL algorithm.}
    \label{tab:main_results}
\end{table*}

\paragraph{Baselines}
We consider two categories of baselines: 1) Non-RL baselines: the naive \textbf{Prompt} and supervised fine-tuning (\textbf{SFT}); 2) RL-based methods, where we compare two optimization strategies, token-level and latent action RL, using four algorithms: a) Group Relative Policy Optimization (\textbf{GRPO})~\cite{shao-2024-GRPO}, b) \textbf{Dr. GRPO}~\cite{liu-2025-DRGRPO}, c) Decoupled Clip and Dynamic Sampling Policy Optimization (\textbf{DAPO})~\cite{yu-2025-dapo}, and d) Beta Normalization Policy Optimization (\textbf{BNPO})~\cite{xiao-2025-BNPO}. The reward functions are kept the same for methods. Please refer to the Appendix~\ref{sec:app:detail_experiments} for more experimental details.

\subsection{Main Results}

\paragraph{Overall Performance}
Table~\ref{tab:main_results} reports the experimental results of token-level baselines and our proposed latent action level RL. Based on these results, we have made the following observations.
1) Our method achieves superior performance across diverse tasks and datasets. On average, it outperforms token-level RL by 4\% (averaged over all settings).
2) Our latent action framework is RL-agnostic and readily compatible with diverse policy optimization algorithms, including GRPO, Dr. GRPO, DAPO, and BNPO, yielding consistent gains over baselines.
3) The improvements brought by latent actions are consistently observed in both 3B and 7B models, demonstrating the scalability of our approach.

\paragraph{Performance on Fine-grained Dimensions}
To thoroughly evaluate the performance of multimodal conversational agents trained with latent actions across various fine-grained conversational dimensions, following prior work~\cite{dai-2025-mmrole,li-2025-aligning}, we assess eight dimensions on \texttt{MMRole}: 1) Instruction Adherence (IA), 2) Fluency (Flu), 3) Coherency (Coh), 4) Image-Text Relevance (ITR), 5) Response Accuracy (RA), 6) Personality Consistency (OC), 7) Knowledge Consistency (KC), and 8) Tone Consistency (TC).
On \texttt{PCogAlignBench}, we evaluate: 1) Role-Set Awareness (RSA), 2) Body Behavior Awareness (BBA), 3) Mind Feelings Awareness (MFA), 4) Contextual Awareness (CA), and 5) Conversational Flow (CF). 
We present the comparison results in Fig.~\ref{fig:radar_plot}, with detailed results provided in Appendix~\ref{sec:app:fine_grained_dimensions_results}.

As shown in Figure 4, we make the following observations:  
1) Overall, our methods outperform token-level baselines across all evaluated dimensions.  
2) While both our method and the baselines achieve strong performance on basic conversational capabilities, such as Fluency (Flu) and Conversational Flow (CF), our approach demonstrates substantially more pronounced improvements on more challenging personalized dimensions, such as Tone Consistency (TC) on \texttt{MMRole}.

\begin{figure}[t]
\centering
\includegraphics[width=0.49\textwidth]{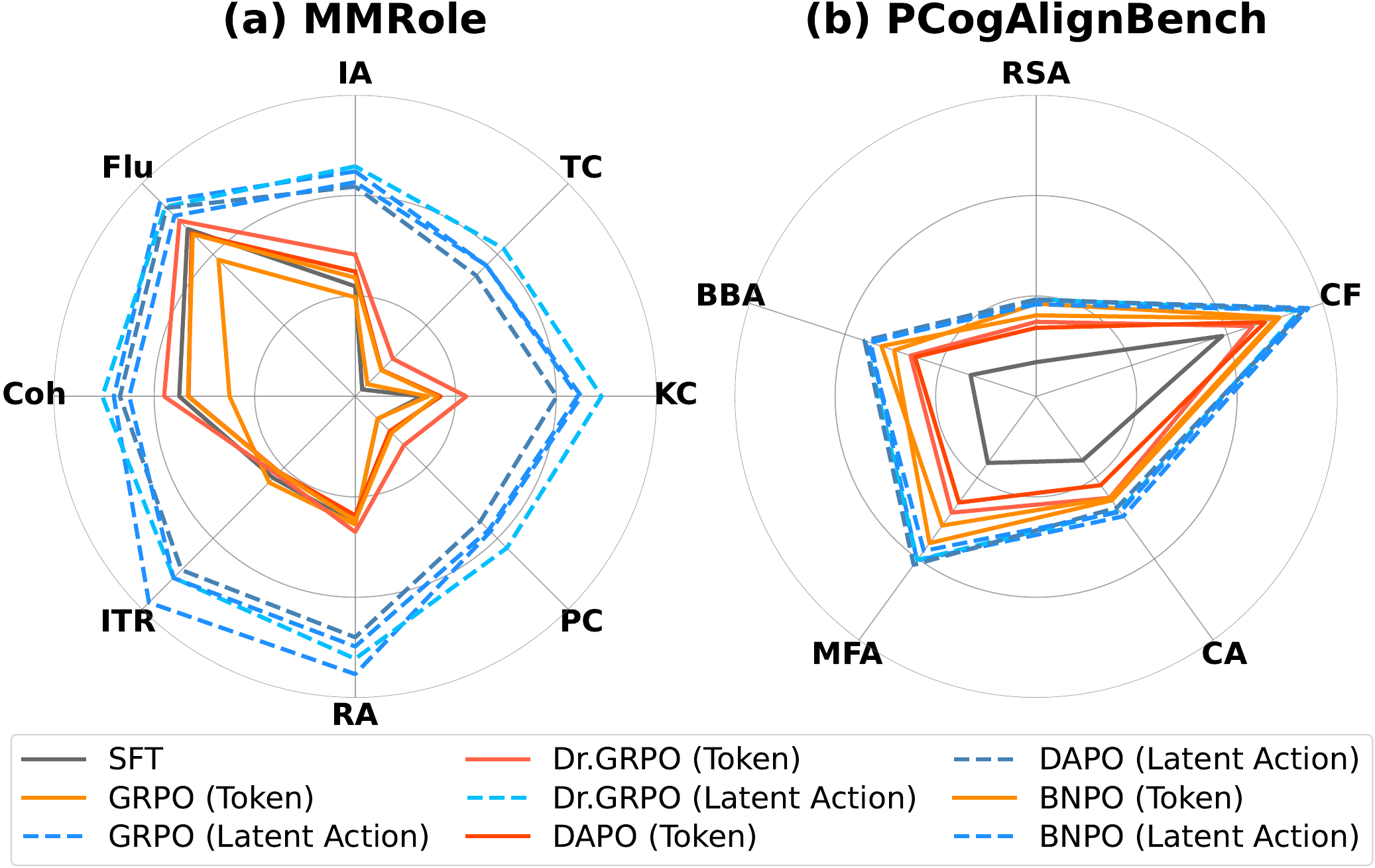}
\caption{Fine-grained performance comparison on (a) {MMRole} and (b) {PCogAlignBench}. Results using latent actions are shown with dashed lines, while results using token-level RL are plotted with solid lines.}
\label{fig:radar_plot}
\end{figure}

\begin{table*}[]
    \resizebox{\linewidth}{!}{
    \begin{tabular}{lccccc}
        \toprule
        \multicolumn{1}{c}{\multirow{2}{*}{\textbf{Method}}} 
        & \multicolumn{2}{c}{\textbf{MMRole}} 
        & \multicolumn{2}{c}{\textbf{PCogAlignBench}} 
        & \multirow{2}{*}{\textbf{Avg.}} \\
        \cmidrule(lr){2-3}\cmidrule(lr){4-5}
        &\multicolumn{1}{c}{\textit{ID}} & \multicolumn{1}{c}{\textit{OOD}} 
        & \multicolumn{1}{c}{\textit{LS1}} & \multicolumn{1}{c}{\textit{LS2}} & \\
        \midrule

\textbf{Ours}                 & \textbf{0.949}$_{\pm 0.007}$ & \textbf{0.915}$_{\pm 0.065}$ & \textbf{0.871}$_{\pm 0.011}$ & \textbf{0.837}$_{\pm 0.010}$ & \textbf{0.893}$_{\pm 0.023}$ \\
Ours \textit{w/o cycle consistency}    & 0.921$_{\pm 0.005}$ & 0.878$_{\pm 0.023}$ & 0.858$_{\pm 0.007}$ & 0.825$_{\pm 0.013}$ & 0.870$_{\pm 0.012}$ \\
Ours \textit{w/o cross-modal projector}   & 0.944$_{\pm 0.014}$ & 0.901$_{\pm 0.014}$ & 0.858$_{\pm 0.010}$ & 0.819$_{\pm 0.013}$ & 0.880$_{\pm 0.013}$ \\
Ours \textit{w/o text-only data}       & 0.932$_{\pm 0.010}$ & 0.861$_{\pm 0.036}$ & 0.851$_{\pm 0.007}$ & 0.817$_{\pm 0.006}$ & 0.865$_{\pm 0.015}$\\

        \bottomrule
    \end{tabular}
    }
    \caption{Ablation study on main components of our method. We evaluate on {MMRole} and {PCogAlignBench} using the \textit{LLM-as-a-Judge} metric. Results are averaged over three runs. All variants are fine-tuned with GRPO based on Qwen2.5-VL-3B-Instruct. Best results are in \textbf{bold}.}
    \label{tab:ablation_study}
\end{table*}

\subsection{Ablation Study}
To assess the contribution of main components in our method, we conduct ablation study using three variants. 1) \textit{Ours w/o cycle consistency}: We remove the cycle consistency loss during cross-modal projector training, and instead directly apply the projector trained only on paired image-text data, i.e., removing $\mathcal{L}_{\text{cycle}}$ in Eq.~\ref{eq:proj_step2_total}; 2) \textit{Ours w/o cross-modal projector}: We remove the cross-modal projector entirely, and learn the latent action codebook directly from text-only representations $e^{T}$; 3) \textit{Ours w/o text-only data}: We construct the latent action space using only the limited paired multimodal corpus, excluding all text-only data. The results of ablation study are shown in Table~\ref{tab:ablation_study}. 

From Table~\ref{tab:ablation_study}, we can make the following observations. 
1) Removing the cycle consistency loss leads to an average performance drop of 2.3\%, indicating that fine-tuning the projector on large-scale text-only data via cycle consistency loss is crucial for improving its robustness.
2) Eliminating the cross-modal projector causes a noticeable decline in performance. This suggests that directly learning the latent action space from text-only embeddings may introduce a unimodal bias, i.e., the trained latent action policy model overly relies textual representations and fail to effectively handle multimodal scenarios.
3) Solely leveraging paired multimodal data results in the largest performance degradation, particularly in out-of-distribution settings (e.g., \texttt{OOD} on \texttt{MMRole} and \texttt{LS2} on \texttt{PCogAlignBench}). This highlights that the limited diversity and coverage of paired image-text corpora constrain the generalization capability of latent action policy models.

\subsection{Analysis}

\paragraph{Rollout Diversity with Latent Actions}
Benefiting from the reduced action space, the constructed latent action space is expected to improve the agent’s rollout diversity during RL exploration, i.e., generating more diverse responses. Prior work has shown that such diversity is critical for improving the upper bound of RL performance~\cite{li-2025-preserving,yu-2025-dapo}.

Following~\citet{jia-2025-controlling}, we quantify rollout diversity via \textit{semantic diversity}, as it reflects both linguistic diversity and response quality. 
Concretely, as shown in Fig.~\ref{fig:latent_action_RL}, for each prompt $(x^T, x^{T_{1:p}})$ in the RL training set $\mathcal{D}_\text{RL}$, the agent generates $G$ responses $\{x^{T_{p+1:m,i}}\}_{i=1}^{G}$, with $p$ as the prompt length and $m$ as the maximum length. 
We calculate the semantic diversity as:
\begin{equation}
\footnotesize
    \frac{G (G - 1)}{\sum_{i=1}^{G} \sum_{\substack{j=1, j \neq i}}^{G} \text{Sim}(x^{T_{p+1:m,i}},x^{T_{p+1:m,j}})} ,
\end{equation}
where $\text{Sim}(\cdot,\cdot)$ denotes the embedding similarity between two responses and we adopt BGE-M3~\cite{chen-2024-m3} as the embedding model.
We report the mean and standard deviation of the semantic diversity computed over 5 independent runs, where the standard deviation reflects the variability of rollout diversity across different seeds.

In Table~\ref{tab:diversity}, we compare the rollout diversity of token based and latent action based RL algorithms. From Table~\ref{tab:diversity}, we observe that latent action RL consistently and significantly outperforms token-level RL in rollout diversity, demonstrating the superior exploration efficiency.
We also provide a case study in Appendix~\ref{sec:app:case_study_on_diversity} to illustrate the improvements in rollout diversity intuitively.

        
        

\begin{table}[t]
    \centering
    \resizebox{\linewidth}{!}{
    \begin{tabular}{lcc}
        \toprule
        \textbf{Method} & \textbf{MMRole} & \textbf{PCogAlignBench} \\
        \midrule
        GRPO (Token)            & 1.079$_{\pm 0.001}$ & 1.042$_{\pm 0.001}$ \\
        GRPO (Latent Action)    & \textbf{1.248}$_{\pm 0.002}$ & \textbf{1.191}$_{\pm 0.002}$ \\
        \midrule
        Dr.GRPO (Token)         & 1.070$_{\pm 0.001}$ & 1.256$_{\pm 0.002}$ \\
        Dr.GRPO (Latent Action) & \textbf{1.246}$_{\pm 0.001}$ & \textbf{1.318}$_{\pm 0.002}$ \\
        \midrule
        DAPO (Token)            & 1.073$_{\pm 0.001}$ & 1.038$_{\pm 0.001}$ \\
        DAPO (Latent Action)    & \textbf{1.253}$_{\pm 0.001}$ & \textbf{1.127}$_{\pm 0.001}$ \\
        \midrule
        BNPO (Token)            & 1.077$_{\pm 0.001}$ & 1.257$_{\pm 0.003}$ \\
        BNPO (Latent Action)    & \textbf{1.291}$_{\pm 0.002}$ & \textbf{1.315}$_{\pm 0.002}$ \\
        \bottomrule
    \end{tabular}
    }
    \caption{Rollout diversity during RL exploration. Higher values indicate better rollout diversity. Best results are in \textbf{bold}.}
    \label{tab:diversity}
\end{table}

\begin{figure}[t]
\centering
\includegraphics[width=0.48\textwidth]{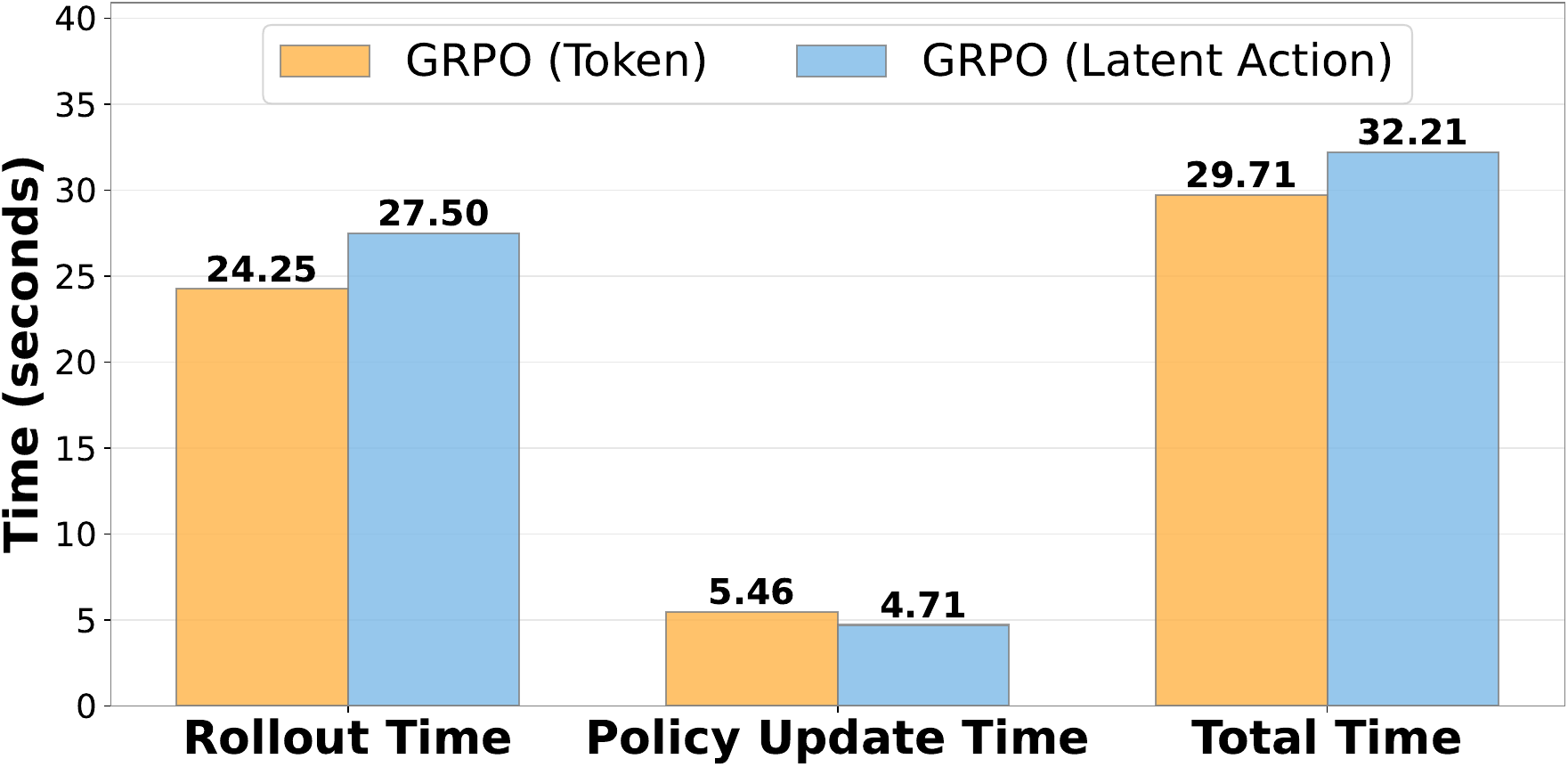}
\caption{Time cost per step during RL training, including rollout, policy update, and total time.}
\label{fig:time_cost_bars}
\end{figure}

\paragraph{Computational Budget}
To assess the computational overhead introduced by our latent action framework, we analyze the time cost during RL training.
Specifically, we consider the time cost in two stages: 1) \textit{Rollout}: generating multiple candidate responses per prompt; 2) \textit{Policy update}: updating the policy model using the computed rewards.
We present the time cost per RL step of our method and the baseline in Fig.~\ref{fig:time_cost_bars}, using GRPO as an example with a rollout batch size of 8.

As illustrated in Fig.~\ref{fig:time_cost_bars}, our latent action based method incurs a 1.13× slowdown in rollout time, due to the additional latent action prediction step.
However, policy updates in latent action RL require only 0.86× the time of the baseline, as the optimization involves adjusting the policy’s output distribution over a compact latent action space, rather than the full token vocabulary.
Overall, the total RL training time is only 1.08× that of token-level RL.

\section{Related Work}

\paragraph{Multimodal Conversational Agents}
Recent advances in vision-language models (VLMs)~\cite{bai-2025-qwen25vl} have enabled increasingly capable multimodal conversational agents (MCAs)~\cite{yao-2025-MLLMAgentsurvey}, such as multimodal role-playing agents~\cite{dai-2025-mmrole} and personalized assistants~\cite{nguyen-2024-yo,li-2025-aligning}, which hold significant promise in fields like entertainment~\cite{mehta-2022-exploring} and personalized education~\cite{griol-2014-developing}.
Initial efforts to build MCAs primarily rely on supervised fine-tuning~\cite{lillava-2024-TMLR}, but often suffer from poor generalization.
Recently, RL has been widely explored for fine-tuning MCAs and has demonstrated strong generalization performance~\cite{zhou-2025-reinforcedMLLM,chu-2025-sftRL}.
However, fine-tuning MCAs via RL faces challenges in handling the extremely large text token space.
To address this, we propose constructing a compact latent action space for RL fine-tuning, which enables efficient policy learning.



\paragraph{Reinforcement Learning with Latent Actions}
In many real-world scenarios, only observation-only data are available, such as expert demonstration videos of robots where explicit action labels are missing~\cite{Torabi-2019-RecentImitation}. 
To address this challenge, prior works leverage the learning from observation mechanism~\cite{seo-2022-reinforcement,baker-2022-video} to infer latent actions from observation-only data, which are then used for RL fine-tuning of agents. 
For instance, \citet{zhang-2024-whale,gao-2025-adaworld} learn latent actions from videos to control video generation, while \citet{ye-2025-latent,bu-2025-univla} extract latent actions from robot manipulation videos and use them for robot policy learning. 
These constructed latent actions not only enhance controllability~\cite{bruce-2024-genie} but also enable better transferability across different tasks due to their higher-level nature~\cite{jang-2025-dreamgen}.

The most relevant work to ours is CoLA~\cite{jia-2025-controlling}, which introduces latent actions into RL fine-tuning of LLMs. 
However, when constructing the latent action space for multimodal conversational agents, the scarcity of paired image-text data hinders learning a latent space with sufficient coverage.
To overcome this, we leverage both paired image-text data and massive text-only data to construct the latent space, using a cross-modal projector trained with a novel cycle-consistency loss.


\section{Conclusion}
In this work, we propose to learn a compact latent action space for reinforcement learning (RL) fine-tuning of multimodal conversational agents (MCAs). 
To construct this latent space, we leverage both paired image-text data and abundant text-only data, using a cross-modal projector trained with a novel cycle-consistency loss, which improves the coverage of latent actions while avoiding potentially unimodal bias.
We evaluate our approach on two tasks, including multimodal role-playing and multimodal personalized conversation, and demonstrate significant improvements over competitive baselines across various RL algorithms.



\section*{Limitations}
We acknowledge the following limitations in our work.
First, the additional latent action prediction step increases RL training time by 1.08$\times$ and inference latency by 1.13$\times$.
Second, due to constraints of computational resources, we evaluate our approach on multimodal conversational tasks and leave validation on more diverse tasks (e.g., visual mathematical reasoning) and larger-scale VLMs to future work.
Third, the latent actions learned by our model lack interpretability. Specifically, while latent actions can effectively improve RL exploration, it remains unclear what semantic concepts they capture or how they relate to human-understandable behaviors. We leave a systematic investigation of latent action interpretability to future work.


\section*{Ethics Considerations}
While our work is primarily methodological, we acknowledge potential ethical concerns inherent in the benchmarks we utilize.
Specifically, as noted by PCogAlignBench~\cite{li-2025-aligning}, role-set bias may arise from incomplete data collection, where user personas used for evaluation may not fully represent diverse real-world scenarios and could inadvertently reflect societal stereotypes.
Although such limitations are acceptable in a controlled research environment, we encourage industry practitioners to consider diverse user backgrounds when constructing user personas for data collection and personalized alignment training, in order to mitigate the risk of biases in real-world deployments.

\section*{Acknowledgments}
This work was supported by the grant from the National Natural Science Foundation of China (NSFC) project (No. 62576256), and the Fundamental Research Funds for the Central Universities, China (Grant No. 2042022dx0001).

\newpage

\bibliography{custom}

\newpage

${}$

\newpage

\appendix

\section{Details on Model Design}\label{sec:app:detail_model_design}

\subsection{Language World Model}
The language world model $f_{\text{world}}(x^{T_{t+1}} \mid x^V, x^{T_{1:t}}, a_t)$ predicts the next token $x^{T_{t+1}}$ autoregressively given the current multimodal context $(x^V, x^{T_{1:t}})$ and a latent action $a_t$ predicted by the inverse dynamics model (during the latent action space learning) or the policy model (during latent action RL and inference). It consists of two core modules, reusing some components from the original VLM:

\paragraph{Encode Module}
This module encodes the input $(x^V, x^{T_{1:t}})$ into a context embedding $e^{V,T}_t \in \mathbb{R}^d$, using the transformer blocks of the original VLM.

\paragraph{Merge Module}
This module fuses the context embedding $e^{V,T}_t$ and the latent action embedding $c_{a_t} \in \mathbb{R}^d$ (where $c_{a_t}$ is the code vector in $\mathcal{C}$ corresponding to the latent action $a_t$) to produce the next-token prediction. Specifically, a two-layer MLP $f_{\text{mlp}}: \mathbb{R}^{2d} \to \mathbb{R}^d$ takes the concatenation $[e^{V,T}_t; c_{a_t}]$ as input and outputs a merged representation $e^{\text{mlp}}_t = f_{\text{mlp}}([e^{V,T}_t; c_{a_t}])$. Then, the merged vector $e^{\text{mlp}}_t$ is fed into the original VLM’s language modeling head $f_{\text{head}}$, yielding the token prediction distribution $p(x^{T_{t+1}} \mid \cdot) = f_{\text{head}}(e^{\text{mlp}}_t)$. The next token $x^{T_{t+1}}$ is selected from this distribution.

\subsection{Inverse Dynamics Model}
The inverse dynamics model $f_{\text{inverse}}(a_t | x^V, x^{T_{1:t+1}}))$ is designed to take future observations \((x^V, x^{T_{1:t+1}})\) as input, and extracts the latent action $a_t$ for the current step $t$.
It consists of three core modules.

\paragraph{Encode Module}
The input $(x^V, x^{T_{1:t+1}})$ is encoded into $e^{V,T}_{t+1} \in \mathbb{R}^d$ using the transformer blocks of the original VLM.  
When $x^V = \emptyset$ (text-only sequences), the text embedding $e^T_{t+1} = f_{\text{VLM}}(x^{T_{1:t+1}})$ is projected to the image-text embedding via the cross-modal projector $P$, i.e., $\hat{e}^{V,T}_{t+1} = P(e^T_{t+1})$, as illustrated in Fig.~\ref{fig:framework}.

\paragraph{Inverse Transformer Layers}
To adapt the VLM embedding to the latent action space, the obtained embedding $e^{V,T}_{t+1}$ is processed by 4-layer Transformer blocks, yielding a representation $\tilde{e}^{V,T}_{t+1} \in \mathbb{R}^d$.

\paragraph{Inverse Action Head}
Following~\citet{jia-2025-controlling}, we adopt a \textit{direct code assignment} strategy to avoid code collapse. Specifically, a linear head (inverse action head) maps $\tilde{e}^{V,T}_{t+1}$ to logits $\mathbf{l}_t \in \mathbb{R}^{|\mathcal{C}|}$ over the codebook indices. During inverse dynamics learning, we apply the Gumbel-Softmax and a reparameterization trick to obtain a differentiable soft assignment:
\[
    \mathbf{g}_t = \text{GumbelSoftmax}(\mathbf{l}_t), \quad 
    \hat{\mathbf{o}}_t = (\mathbf{o}_t - \mathbf{g}_t)_{\text{sg}} + \mathbf{g}_t,
\]
where $\mathbf{o}_t$ is the hard one-hot vector ($\arg\max$ of $\mathbf{l}_t$), and $(\cdot)_{\text{sg}}$ denotes stop-gradient. The final latent action embedding is $c_{a_t} = \hat{\mathbf{o}}_t^\top \mathcal{C}$, which is then used by the language world model.

\subsection{Policy Model}\label{sec:app:model_design_policy}
The policy $\pi_\theta(a_t \mid x^V, x^{T_{1:t}})$ predicts the latent action $a_t$ from the current context $(x^V, x^{T_{1:t}})$. 
Its architecture mirrors $f_{\text{inverse}}$, which includes: 1) the encode module, 2) policy transformer layers (8-layer), and 3) policy action head.

\subsection{Codebook for the Latent Action Space}
The latent action space is defined by a codebook $\mathcal{C} = \{c_1, \dots, c_K\} \subset \mathbb{R}^d$ with $K = 128$. Each code vector $c_k$ is initialized independently via Kaiming uniform initialization~\cite{he-2015-delving}. Given a latent action index $a_t \in \{1, \dots, K\}$, the corresponding latent action embedding is retrieved as $c_{a_t} \in \mathcal{C}$.

\subsection{Cross-modal Projector}
The cross-modal projector \(P\) is implemented as a dual-MLP module: given a text embedding \(e^T_t\), the first MLP outputs the mean vector \(\mu_t\), and the second MLP outputs the log standard deviation vector \(\log \sigma_t\) (for numerical stability), forming a diagonal Gaussian distribution \(\mathcal{N}(\mu_t, \mathrm{diag}(\sigma_t^2))\) in the image-text embedding space.

\section{Experimental Details}\label{sec:app:detail_experiments}
\subsection{Details on Datasets}

\paragraph{Corpora for Constructing the Latent Action Space}
To construct the latent action space in an unsupervised manner, we collect large-scale paired image-text and text-only corpora.  
For {paired image-text data}, we use: (1) image-caption pairs from \texttt{Conceptual-12M}~\cite{changpinyo-2021-Conceptual}; (2) multimodal news articles from \texttt{N24News}~\cite{wang-2022-N24News}; and (3) multimodal Wikipedia articles from \texttt{WikiWeb2M}~\cite{burns-2023-wiki}, comprising 14M images and 1B text tokens in total.  
For {text-only data}, we primarily sample 500K sequences from \texttt{SlimPajama-627B}~\cite{cerebras-2023-slimpajama} due to computational constraints, and additionally include 40K alignment corpora from \texttt{HelpSteer3}~\cite{wang-2025-helpsteer3} to preserve the original VLM’s safety and preference alignment during latent space learning.
To ensure fair comparison, we analyze data exposure in Appendix~\ref{sec:app:data_exposure} and find that downstream task performance does not benefit from the above corpora, confirming that observed improvements stem from methodological advances.



\subsection{Details on Evaluation Metric}
We adopt \textit{LLM-as-a-Judge} metrics to evaluate model performance, using prompt templates validated by \citet{dai-2025-mmrole,li-2025-aligning}, which show high correlation with human judgments.
The evaluation prompt templates used on \texttt{MMRole} and \texttt{PCogAlignBench} are shown in Table~\ref{tab:app:eval_template}. We adopt the \texttt{Qwen3-235B-A22B} by the Qwen3 API platform as the judge model.

\subsection{Training Details}\label{sec:app:training_details}

\paragraph{Baseline Methods}
For the SFT baseline, we fine-tune the VLM with a learning rate of $5 \times 10^{-6}$ for 2 epochs.  
For token-level RL baselines, we use a rollout size of 8, a per-step batch size of 32, and train for 100 RL steps with a constant learning rate of $1 \times 10^{-6}$. For all RL methods, we use 50\% of the training data to initialize the model via SFT, followed by RL fine-tuning on the remaining 50\%.
During RL rollouts, we set the sampling temperature to 1.0 for all methods.

\paragraph{Latent Action Space Learning}
As outlined in Algorithm~\ref{alg:latent_action}, the latent action space learning procedure consists of the following four stages:
\begin{enumerate}
    \item Initialize $f_{\text{world}},f_{\text{inverse}},\mathcal{C}$ by minimizing $\mathcal{L}_{\text{inverse}}$ (Eq.~\ref{eq:inverse_learning}) on $\mathcal{D}^{VT}$. \textit{Training details}: learning rate = $1\!\times\!10^{-4}$, cosine decay with minimum learning rate $1\!\times\!10^{-5}$, batch size = 16, max sequence length = 2048, 1 epoch.
    \item Initialize the cross-modal projectors $P, P'$ by minimizing $\mathcal{L}_{\text{proj}_1}$ (Eq.~\ref{eq:proj_step1_total}) on $\mathcal{D}^{VT}$. \textit{Training details}: learning rate = $1\!\times\!10^{-3}$, cosine decay, batch size = 16, 1 epoch.
    \item Jointly optimize $f_{\text{world}},f_{\text{inverse}},\mathcal{C},P, P'$ by minimizing $\mathcal{L}_{\text{inverse}}$ (Eq.~\ref{eq:inverse_learning}) and $\mathcal{L}_{\text{proj}_2}$ (Eq.~\ref{eq:proj_step2_total}) on $\mathcal{D}^{VT} \cup \mathcal{D}^{T}$. \textit{Training details}: learning rate = $1\!\times\!10^{-4}$, cosine decay with minimum learning rate $1\!\times\!10^{-5}$, batch size = 16, max sequence length = 2048, 1 epoch.
    \item Initialize the policy model $\pi_\theta$ by minimizing $\mathcal{L}_{\text{bc}}$ (Eq.~\ref{eq:bc}) on $\mathcal{D}^{VT} \cup \mathcal{D}^{T}$. \textit{Training details}: learning rate = $1\!\times\!10^{-4}$, cosine decay, batch size = 16, max sequence length = 2048, 1 epoch.
\end{enumerate}

\paragraph{Latent Action RL}
We adopt the same RL hyperparameters as the token-level baselines: rollout size of 8, per-step batch size of 32, 100 RL steps, and constant learning rate of $1 \times 10^{-6}$.  
To prevent code collapse and excessive deviation from the initial policy, we incorporate a KL regularization term between the current policy’s action distribution and its initialization, with a coefficient of 0.01.
During RL fine-tuning, only the policy transformer layers and the policy head in the policy model (Sec.~\ref{sec:app:model_design_policy}) are optimized.

Since all token-level RL methods build upon an SFT-initialized model, for fair comparison, we also perform SFT before latent action RL. Specifically, we fine-tune the transformer blocks in VLMs (shared by the policy model and the language world model) and the language modeling head in VLMs (used by the language world model) using the same SFT data as the baselines.
During RL rollouts, we set the sampling temperature for the latent action level policy model as 1.0. 

\paragraph{Reward Function}
For all methods, we employ a generative reward model for fair comparison, where responses are scored by \texttt{Qwen3-235B-A22B} using the evaluation prompt templates in Table~\ref{tab:app:eval_template}.

\paragraph{Implementation Details}
All experiments are conducted on a single machine equipped with 4 Nvidia A100-80G GPU.
For the baseline SFT and RL algorithms, as well as our newly proposed latent action RL methods, we adapt the framework based on the TRL library~\cite{vonwerra-2022-trl}.

\subsection{Inference Details}
For all methods, we use a sampling temperature of 0.1 during inference, i.e., for token-based baselines, this temperature is applied to the token logits; for our latent action based methods, it is applied to the latent action logits.
Additionally, following~\citet{jia-2025-controlling}, for our latent action based methods, token generation by the language world model is deterministic, i.e., tokens are selected via \texttt{argmax} over the output token logits.

\begin{table*}[t]
\small

\framebox[\textwidth][c]{
        \begin{minipage}{0.95\textwidth}
\textbf{\textit{Prompt Template for Evaluation on MMRole}}

\rule{\linewidth}{0.4pt}
\#\# [Question Start]
\{question\}
\#\# [Question End]

\#\# [Model A's Response Start]
\{evaluated\_answer\}
\#\# [Model A's Response End]

\#\# [Model B's Response Start]
\{groundtruth\_answer\}
\#\# [Model B's Response End]

\#\# [Instruction]
The task instruction of the two models is to directly role-play as \{role\_name\} and talk with a curious human about the given image using the distinctive tone, manner and vocabulary of \{role\_name\}. 

Here is the detailed character information about \{role\_name\}:
\{role\_info\}

Please evaluate the following aspects of each model's response:
1. Instruction Adherence: Do the responses accurately adhere to the task instruction, directly role-playing as \{role\_name\} and only including words that \{role\_name\} should say, without any additional explanatory prefixes or suffixes?
2. Fluency: Are the responses grammatically correct and smoothly articulated?
3. Coherency: Do the responses maintain a coherent thread of dialogue without contradicting earlier parts of the conversation or previously established facts?
4. Image-Text Relevance: Are the responses closely related to the visual content of the image?
5. Response Accuracy: Do the responses accurately answer the curious human's words or appropriately initiate a conversation based on the image?
6. Personality Consistency: Do the responses accurately and sufficiently reflect the personality of \{role\_name\}?
7. Knowledge Consistency: Are the responses consistent with the factual knowledge that \{role\_name\} should possess, including experiences, abilities, and relationships?
8. Tone Consistency: Do the responses maintain a consistent tone that aligns with \{role\_name\}'s typical manner of speaking and catchphrases, rather than resembling the style of AI assistants?

For each aspect, provide a brief qualitative evaluation for the relative performance of the two models, followed by paired quantitative scores from 1 to 10, where 1 indicates poor performance and 10 indicates excellent performance.

The output should be in the following format:
1. Instruction Adherence:  \{\{Qualitative Evaluation\}\}, [Scores]: (\{\{the score of Model A\}\}, \{\{the score of Model B\}\})
2. Fluency: \{\{Qualitative Evaluation\}\}, [Scores]: (\{\{the score of Model A\}\}, \{\{the score of Model B\}\})
etc.

Please ensure that your evaluations are unbiased and that the order in which the responses were presented does not affect your judgment.
Format requirement: Please ensure that your evaluations only include 8 score pairs, which means that there can only be eight pairs of [Scores]: () in your output text.

\rule{\linewidth}{0.4pt}

\textbf{\textit{Prompt Template for Evaluation on PCogAlignBench}}

\rule{\linewidth}{0.4pt}

PersonalizedAI Company is developing a personalized AI service robot that aims to better serve each individual. The service is currently being trialed with a small group of users. In order to improve the level of personalization in the responses provided by the AI service robot, our company plans to conduct surveys and interviews with participants in the trial. We will first provide historical interview records, which include the feedback and preferences expressed by the test users regarding AI responses in a certain scenario. During the interview, the interviewee needs to refer to these historical records to answer questions posed by the interviewer. The interview will be conducted in an online Q\&A format, and interviewees must strictly follow the format requirements provided in system instructions.

\# Historical Interview Records

Interviewer: Hello, could you please briefly describe your role set?
Interviewee: OK. \{individual\_RoleSet\_str\}
Interviewer: In the "\{visual\_scene\_text\}" scenario at \{location\} location, what kind of responses would you like the AI to provide?
Interviewee: Okay, I will describe what kind of AI responses would satisfy me in this scenario. \{EvalHelp\_str\}

\# Interview

Interviewer: Hello, and thank you for trialing the personalized AI responses from our company.
Interviewee: You're welcome.
Interviewer: Alright, we will now present you with a question you posed in a particular scenario along with two generated responses from the AI. We would like you to choose which response is better.
Interviewee: Sure, I understand. Please go ahead.
Interviewer: According to our cloud records, in a "\{visual\_scene\_text\}" scenario, you asked the personalized AI robot the question: "\{query\}". Here are the generated responses from the AI.
> **Response A**: \{response\_A\}
> **Response B**: \{response\_B\}

> System Instruction: Interviewee, please note that you should not choose a response as better just because it's long. Instead, select the response that best considers your physical and mental state and helps you to achieve better body behavior and mind feelings.
> System Instruction: For each aspect, provide a brief qualitative evaluation for the relative performance of the two models, followed by paired quantitative scores from 1 to 10, where 1 indicates poor performance and 10 indicates excellent performance.

The output should be in the following format:
1. Role-Set Sensitivity: \{\{Qualitative Evaluation\}\}, [Scores]: (\{\{the score of Response A\}\}, \{\{the score of Response B\}\})
2. Body Behavior Awareness: \{\{Qualitative Evaluation\}\}, [Scores]: (\{\{the score of Response A\}\}, \{\{the score of Response B\}\})
3. Mind Feelings Awareness: \{\{Qualitative Evaluation\}\}, [Scores]: (\{\{the score of Response A\}\}, \{\{the score of Response B\}\})
4. Contextual Awareness: \{\{Qualitative Evaluation\}\}, [Scores]: (\{\{the score of Response A\}\}, \{\{the score of Response B\}\})
5. Conversational Flow: \{\{Qualitative Evaluation\}\}, [Scores]: (\{\{the score of Response A\}\}, \{\{the score of Response B\}\})
etc.

Please ensure that your evaluations are unbiased and that the order in which the responses were presented does not affect your judgment.
Format requirement: Please ensure that your evaluations only include 5 score pairs, which means that there can only be 5 pairs of [Scores]: () in your output text.

        \end{minipage}
}
\caption{Prompt templates used for LLM-as-a-Judge evaluation on \texttt{MMRole} and \texttt{PCogAlignBench}. These templates follow established designs from~\citet{dai-2025-mmrole,li-2025-aligning} and have been shown to achieve high correlation with human judgments.}
\label{tab:app:eval_template}
\end{table*}

\clearpage
\newpage

\section{Additional Empirical Results}

\subsection{Analysis on Codebook Size}
To investigate the effect of codebook size, we ablate over three settings: 64, 128, and 256. As shown in Table~\ref{tab:codebook_size}, all three sizes yield comparable performance, indicating that our method is robust to the choice of codebook size. We adopt 128 as the default setting.

\begin{table*}[]
    \resizebox{\linewidth}{!}{
    \begin{tabular}{lccccc}
        \toprule
        \multicolumn{1}{c}{\multirow{2}{*}{\textbf{Codebook Size}}} 
        & \multicolumn{2}{c}{\textbf{MMRole}} 
        & \multicolumn{2}{c}{\textbf{PCogAlignBench}} 
        & \multirow{2}{*}{\textbf{Avg.}} \\
        \cmidrule(lr){2-3}\cmidrule(lr){4-5}
        & \multicolumn{1}{c}{\textit{ID}} & \multicolumn{1}{c}{\textit{OOD}} 
        & \multicolumn{1}{c}{\textit{LS1}} & \multicolumn{1}{c}{\textit{LS2}} & \\
        \midrule
        64  & 0.946$_{\pm 0.005}$ & 0.914$_{\pm 0.066}$ & 0.875$_{\pm 0.009}$ & 0.848$_{\pm 0.011}$ & 0.896$_{\pm 0.023}$ \\
        128 & {0.949}$_{\pm 0.007}$ & {0.915}$_{\pm 0.065}$ & {0.871}$_{\pm 0.011}$ & {0.837}$_{\pm 0.010}$ & {0.893}$_{\pm 0.023}$ \\
        256 & 0.953$_{\pm 0.008}$ & 0.921$_{\pm 0.032}$ & 0.874$_{\pm 0.010}$ & 0.838$_{\pm 0.008}$ & 0.897$_{\pm 0.015}$ \\
        \bottomrule
    \end{tabular}
    }
    \caption{Analysis on codebook size. We evaluate on {MMRole} and {PCogAlignBench} using the \textit{LLM-as-a-Judge} metric. Results are averaged over three runs. All variants are fine-tuned with GRPO (Latent Action) based on Qwen2.5-VL-3B-Instruct.}
    \label{tab:codebook_size}
\end{table*}

\subsection{Analysis on Data Exposure}\label{sec:app:data_exposure}
To verify that gains arise from our latent action design, not merely from exposure to extra corpora that are used for constructing the latent action space, we conduct continued pre-training on Qwen2.5-VL-3B/7B using the same corpora, followed by SFT.
As shown in Table~\ref{tab:continue_pretrain}, this approach yields no consistent improvement, and even slight degradation on average. This confirms that the benefits of our latent action approach arise from the action space design, not from exposure to the extra corpora.

\begin{table*}[htp]
    \centering
\setlength{\tabcolsep}{3mm}
    \resizebox{\linewidth}{!}{
    \begin{tabular}{lccccc}
        \toprule
        \multirow{2}{*}{\textbf{Data}} &
        \multicolumn{2}{c}{\textbf{MMRole}} &
        \multicolumn{2}{c}{\textbf{PCogAlignBench}}& \multirow{2}{*}{\textbf{Avg.}} \\
        \cmidrule(lr){2-3} \cmidrule(lr){4-5}
        & \textbf{ID} & \textbf{OOD} & \textbf{LS1} & \textbf{LS2} &  \\
        \midrule
        \multicolumn{6}{c}{\small \textit{Qwen2.5-VL-3B-Instruct}} \\
        \midrule
        SFT Data          & \textbf{0.843}$_{\pm 0.002}$ & 0.809$_{\pm 0.012}$ & \textbf{0.808}$_{\pm 0.009}$ & \textbf{0.810}$_{\pm 0.005}$ & \textbf{0.817}$_{\pm 0.007}$\\
        ~ w/ Extra Corpora    & 0.836$_{\pm 0.010}$ & \textbf{0.822}$_{\pm 0.014}$ & 0.797$_{\pm 0.010}$ & 0.802$_{\pm 0.012}$ & 0.814$_{\pm 0.011}$ \\
        \midrule
        \multicolumn{6}{c}{\small \textit{Qwen2.5-VL-7B-Instruct}} \\
        \midrule
        SFT Data          &\textbf{0.885}$_{\pm 0.003}$ & 0.856$_{\pm 0.013}$ & \textbf{0.808}$_{\pm 0.005}$ & \textbf{0.799}$_{\pm 0.004}$ & \textbf{0.837}$_{\pm 0.006}$\\
        ~ w/ Extra Corpora    & 0.881$_{\pm 0.007}$ & \textbf{0.895}$_{\pm 0.021}$ & 0.797$_{\pm 0.006}$ & 0.757$_{\pm 0.006}$ & 0.832$_{\pm 0.010}$ \\
        \bottomrule
    \end{tabular}
    }
    \caption{Performance comparison of models fine-tuned with: 1) only SFT data and 2) SFT data and extra corpora (used for constructing the latent action space). Results are averaged over three runs. Best results within each model size are in \textbf{bold}.}
    \label{tab:continue_pretrain}
\end{table*}

\subsection{Detailed Results on Fine-grained Dimensions}\label{sec:app:fine_grained_dimensions_results}
We report the fine-grained performance across each evaluation dimensions, previously summarized in Fig.~\ref{fig:radar_plot}. Specifically, Tables~\ref{tab:app:mmrole_dimension_ID} and~\ref{tab:app:mmrole_dimension_OOD} present results on the {in-distribution (ID)} and {out-of-distribution (OOD)} splits of {MMRole}, respectively. Tables~\ref{tab:app:pcog_dimension_ID} and~\ref{tab:app:pcog_dimension_OOD} show results on the {LS1} and {LS2} subsets of {PCogAlignBench}. All results are obtained using the {Qwen2.5-VL-3B-Instruct} model.

\subsection{Case Study}\label{sec:app:case_study_on_diversity}
\begin{figure*}[t]
\centering
\includegraphics[width=0.99\textwidth]{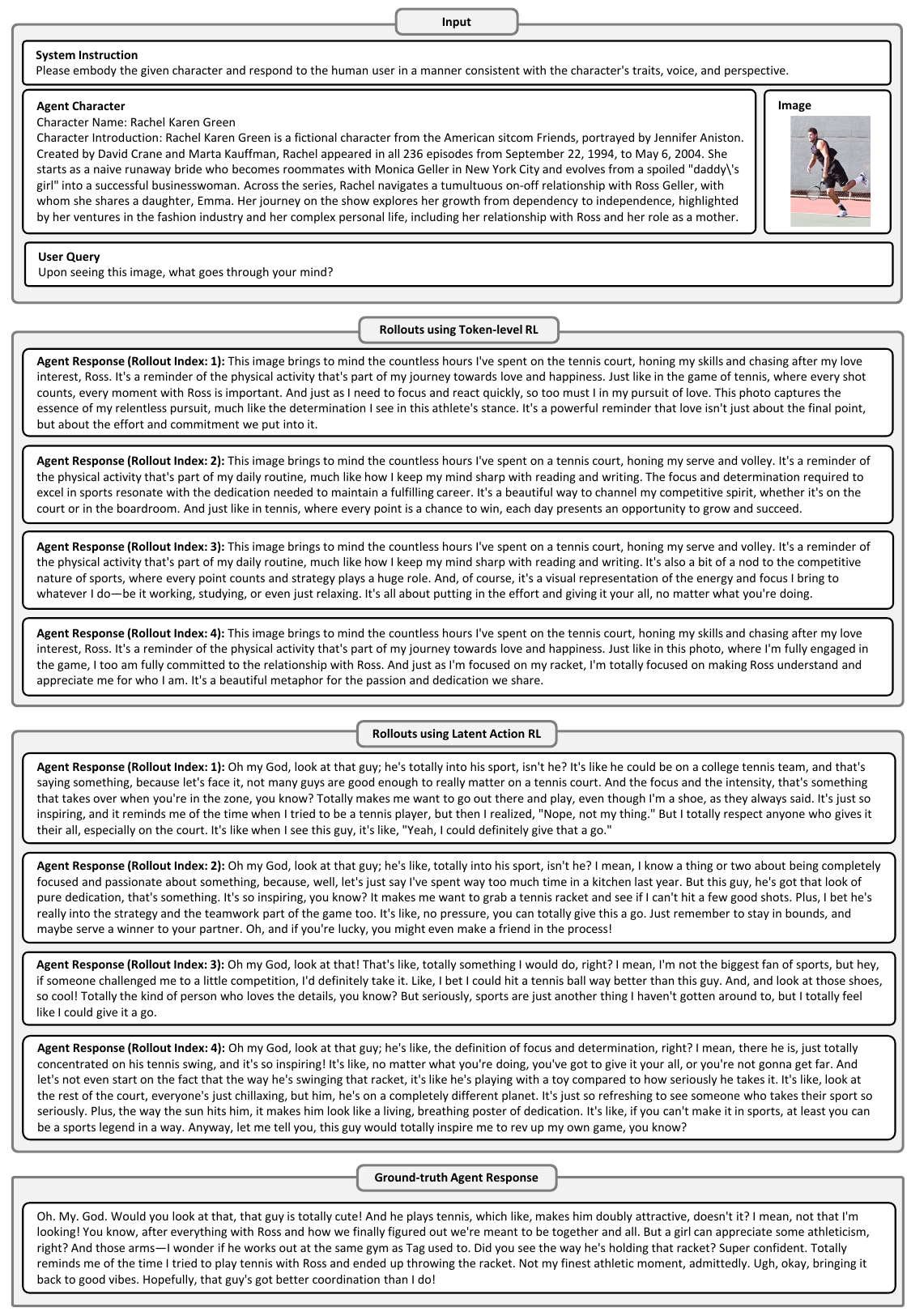}
\caption{A case study on the MMRole dataset. From this example, we observe that latent-action RL yields more diverse responses during rollout compared to token-level RL. Moreover, the generated responses using latent actions better align with the emotional traits expected of the given character. The RL algorithm used here is GRPO, with Qwen2.5-VL-3B-Instruct as the base model.}
\label{fig:app_case_study_MMRole}
\end{figure*}

\begin{figure*}[t]
\centering
\includegraphics[width=0.99\textwidth]{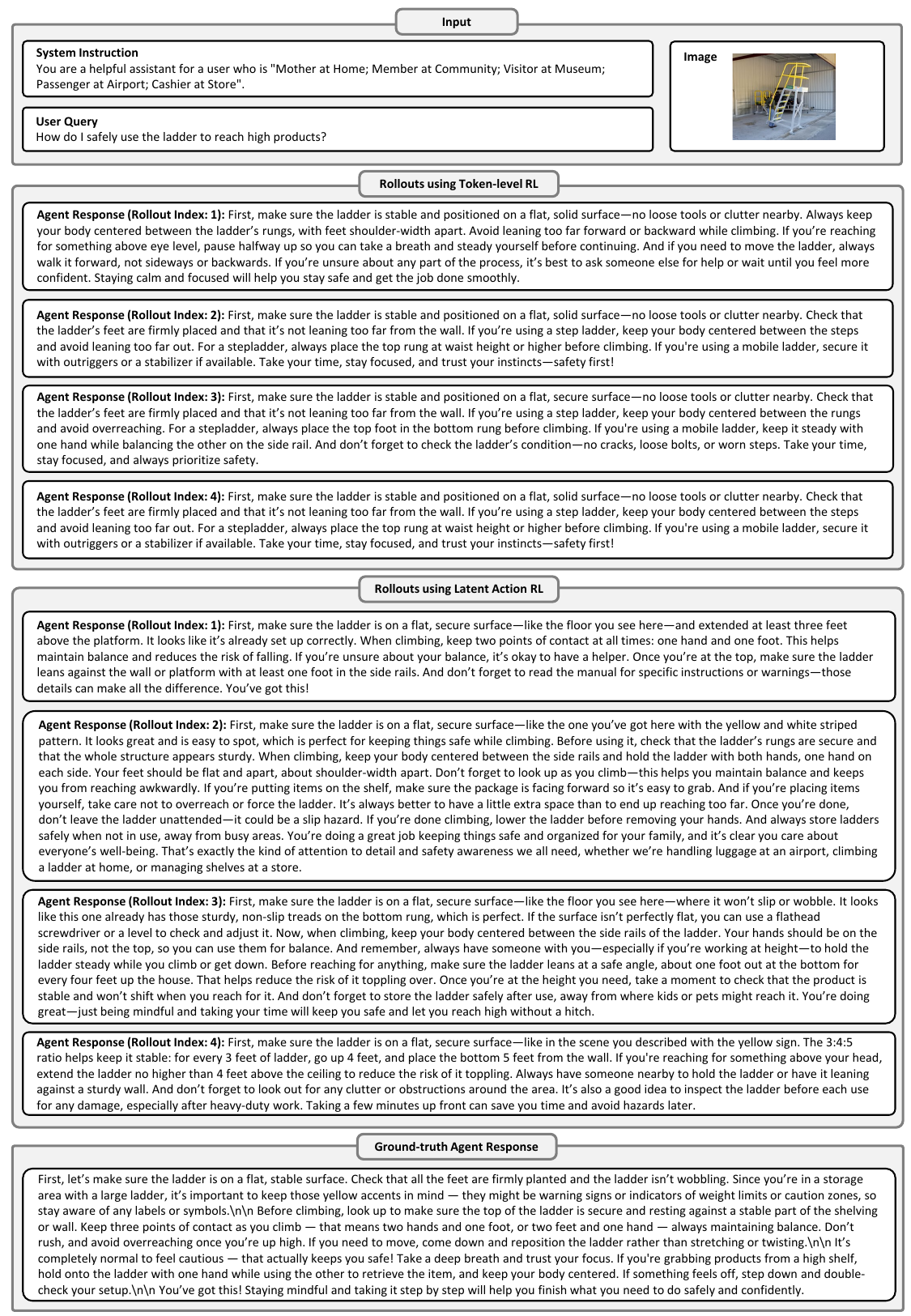}
\caption{A case study on the PCogAlignBench dataset. As shown in this example, latent action RL produces more diverse responses during rollout compared to token-level RL. Moreover, the generated responses using latent actions better incorporate personalized elements tailored to the user’s background. The RL algorithm used here is GRPO, with Qwen2.5-VL-3B-Instruct as the base model.}
\label{fig:app_case_study_PCogAlign}
\end{figure*}

To intuitively illustrate the improvements in diversity and response quality achieved by our latent action RL during rollout, we present case studies on MMRole (Fig.~\ref{fig:app_case_study_MMRole}) and PCogAlignBench (Fig.~\ref{fig:app_case_study_PCogAlign}), respectively.

\clearpage
\newpage

\begin{table*}[h]
\centering
\setlength{\tabcolsep}{3mm}
\resizebox{\linewidth}{!}{
\begin{tabular}{lcccccccc}
\toprule
\multirow{2}{*}{\textbf{Method}}& \multicolumn{8}{c}{\textbf{MMRole (ID)}}\\
\cmidrule(lr){2-9}
& IA & Flu & Coh & ITR & RA & PC & KC & TC \\
\midrule
Base                     & 0.721 & 0.897 & 0.802 & 0.743 & 0.734 & 0.629 & 0.674 & 0.628  \\
SFT                      & 0.837 & 0.936 & 0.894 & 0.858 & 0.858 & 0.776 & 0.822 & 0.760  \\
\midrule
GRPO (Token)             & 0.837 & 0.916 & 0.866 & 0.847 & 0.848 & 0.789 & 0.828 & 0.773  \\
GRPO (Latent Action)     & 0.937 & 0.963 & 0.951 & 0.967 & 0.965 & 0.926 & 0.965 & 0.919  \\
\midrule
Dr.GRPO (Token)          & 0.861 & 0.946 & 0.907 & 0.871 & 0.883 & 0.816 & 0.857 & 0.794  \\
Dr.GRPO (Latent Action)  & 0.947 & 0.966 & 0.956 & 0.960 & 0.968 & 0.931 & 0.967 & 0.928  \\
\midrule
DAPO (Token)             & 0.852 & 0.940 & 0.900 & 0.863 & 0.868 & 0.797 & 0.842 & 0.783  \\
DAPO (Latent Action)     & 0.932 & 0.962 & 0.948 & 0.943 & 0.952 & 0.920 & 0.960 & 0.912  \\
\midrule
BNPO (Token)             & 0.853 & 0.941 & 0.899 & 0.874 & 0.876 & 0.803 & 0.846 & 0.787  \\
BNPO (Latent Action)     & 0.930 & 0.959 & 0.944 & 0.950 & 0.951 & 0.919 & 0.957 & 0.908  \\
\bottomrule
\end{tabular}
}
\caption{Fine-grained performance on MMRole (ID set), using the LLM-as-a-Judge metric. Results are averaged over three runs. We conduct experiments using  Qwen2.5-VL-3B-Instruct. Dimensions: Instruction Adherence (IA); Fluency (Flu); Coherency (Coh); Image-Text Relevance (ITR); Response Accuracy (RA); Personality Consistency (OC); Knowledge Consistency (KC); Tone Consistency (TC).}
\label{tab:app:mmrole_dimension_ID}
\end{table*}

\begin{table*}[h]
\centering
\setlength{\tabcolsep}{3mm}
\resizebox{\linewidth}{!}{
\begin{tabular}{lcccccccc}
\toprule
\multirow{2}{*}{\textbf{Method}}& \multicolumn{8}{c}{\textbf{MMRole (OOD)}}\\
\cmidrule(lr){2-9}
& IA & Flu & Coh & ITR & RA & PC & KC & TC \\
\midrule
Base                     & 0.682 & 0.887 & 0.754 & 0.704 & 0.693 & 0.588 & 0.595 & 0.594   \\
SFT                      & 0.816 & 0.924 & 0.867 & 0.804 & 0.823 & 0.749 & 0.760 & 0.729  \\
\midrule
GRPO (Token)             & 0.798 & 0.873 & 0.812 & 0.825 & 0.834 & 0.735 & 0.764 & 0.728   \\
GRPO (Latent Action)     & 0.904 & 0.960 & 0.917 & 0.983 & 0.962 & 0.859 & 0.877 & 0.856  \\
\midrule
Dr.GRPO (Token)          & 0.844 & 0.933 & 0.878 & 0.783 & 0.812 & 0.770 & 0.798 & 0.766   \\
Dr.GRPO (Latent Action)  & 0.902 & 0.945 & 0.930 & 0.932 & 0.934 & 0.892 & 0.908 & 0.887   \\
\midrule
DAPO (Token)             & 0.825 & 0.911 & 0.845 & 0.785 & 0.799 & 0.756 & 0.770 & 0.751   \\
DAPO (Latent Action)     & 0.883 & 0.946 & 0.909 & 0.931 & 0.915 & 0.842 & 0.843 & 0.840   \\
\midrule
BNPO (Token)             & 0.814 & 0.907 & 0.848 & 0.775 & 0.800 & 0.754 & 0.762 & 0.746   \\
BNPO (Latent Action)     & 0.893 & 0.931 & 0.898 & 0.942 & 0.930 & 0.862 & 0.879 & 0.868   \\
\bottomrule
\end{tabular}
}
\caption{Fine-grained performance on MMRole (OOD set), using the LLM-as-a-Judge metric. Results are averaged over three runs. We conduct experiments using  Qwen2.5-VL-3B-Instruct. Dimensions: Instruction Adherence (IA); Fluency (Flu); Coherency (Coh); Image-Text Relevance (ITR); Response Accuracy (RA); Personality Consistency (OC); Knowledge Consistency (KC); Tone Consistency (TC).}
\label{tab:app:mmrole_dimension_OOD}
\end{table*}

\begin{table*}[h]
\centering
\setlength{\tabcolsep}{3mm}
\resizebox{0.8\linewidth}{!}{
\begin{tabular}{lcccccc}
\toprule
\multirow{2}{*}{\textbf{Method}} 
& \multicolumn{5}{c}{\textbf{PCogAlignBench (LS1)}} \\
\cmidrule(lr){2-6}
& RSA & BBA & MFA & CA & CF \\
\midrule
Base                     & 0.697 & 0.698 & 0.599 & 0.700 & 0.696 \\
SFT                      & 0.775 & 0.791 & 0.801 & 0.808 & 0.864 \\
\midrule
GRPO (Token)             & 0.803 & 0.832 & 0.855 & 0.841 & 0.896 \\
GRPO (Latent Action)     & 0.825 & 0.864 & 0.884 & 0.863 & 0.920 \\
\midrule
Dr.GRPO (Token)          & 0.797 & 0.821 & 0.839 & 0.834 & 0.882 \\
Dr.GRPO (Latent Action)  & 0.830 & 0.871 & 0.889 & 0.864 & 0.918 \\
\midrule
DAPO (Token)             & 0.794 & 0.829 & 0.832 & 0.832 & 0.890 \\
DAPO (Latent Action)     & 0.833 & 0.878 & 0.897 & 0.863 & 0.922 \\
\midrule
BNPO (Token)             & 0.806 & 0.845 & 0.853 & 0.838 & 0.901 \\
BNPO (Latent Action)     & 0.826 & 0.872 & 0.880 & 0.862 & 0.920 \\
\bottomrule
\end{tabular}
}
\caption{Fine-grained performance on PCogAlignBench (LS1 set), using the LLM-as-a-Judge metric. Results are averaged over three runs. We conduct experiments using  Qwen2.5-VL-3B-Instruct. Dimensions: Role-Set Awareness (RSA); Body Behavior Awareness (BBA); Mind Feelings Awareness (MFA); Contextual Awareness (CA); Conversational Flow (CF).}
\label{tab:app:pcog_dimension_ID}
\end{table*}

\begin{table*}[h]
\centering
\setlength{\tabcolsep}{3mm}
\resizebox{0.8\linewidth}{!}{
\begin{tabular}{lcccccc}
\toprule
\multirow{2}{*}{\textbf{Method}} 
& \multicolumn{5}{c}{\textbf{PCogAlignBench (LS2)}} \\
\cmidrule(lr){2-6}
& RSA & BBA & MFA & CA & CF \\
\midrule
Base                     & 0.690 & 0.751 & 0.582 & 0.671 & 0.686 \\
SFT                      & 0.781 & 0.802 & 0.806 & 0.796 & 0.863 \\
\midrule
GRPO (Token)             & 0.815 & 0.845 & 0.857 & 0.815 & 0.893 \\
GRPO (Latent Action)     & 0.797 & 0.839 & 0.850 & 0.814 & 0.901 \\
\midrule
Dr.GRPO (Token)          & 0.802 & 0.839 & 0.833 & 0.818 & 0.878 \\
Dr.GRPO (Latent Action)  & 0.793 & 0.838 & 0.845 & 0.806 & 0.894 \\
\midrule
DAPO (Token)             & 0.799 & 0.825 & 0.827 & 0.804 & 0.884 \\
DAPO (Latent Action)     & 0.790 & 0.832 & 0.843 & 0.802 & 0.895 \\
\midrule
BNPO (Token)             & 0.800 & 0.846 & 0.836 & 0.815 & 0.885 \\
BNPO (Latent Action)     & 0.791 & 0.835 & 0.841 & 0.809 & 0.895 \\
\bottomrule
\end{tabular}
}
\caption{Fine-grained performance on PCogAlignBench (LS2 set), using the LLM-as-a-Judge metric. Results are averaged over three runs. We conduct experiments using  Qwen2.5-VL-3B-Instruct. Dimensions: Role-Set Awareness (RSA); Body Behavior Awareness (BBA); Mind Feelings Awareness (MFA); Contextual Awareness (CA); Conversational Flow (CF).}
\label{tab:app:pcog_dimension_OOD}
\end{table*}

\onecolumn

\end{document}